\documentclass[a4paper,conference]{IEEEtran}
\usepackage{ifpdf}
\usepackage{cite}
\ifCLASSINFOpdf
   \usepackage[pdftex]{graphicx}
   \DeclareGraphicsExtensions{.pdf,.jpeg,.png}
\else
   \usepackage[dvips]{graphicx}
   \DeclareGraphicsExtensions{.eps}
\fi

\usepackage{amsmath}
\usepackage{algorithmic}
\usepackage{array}

\ifCLASSOPTIONcompsoc
  \usepackage[caption=false,font=normalsize,labelfont=sf,textfont=sf]{subfig}
\else
  \usepackage[caption=false,font=footnotesize]{subfig}
\fi
\usepackage{stfloats}

\usepackage{authblk}
\usepackage{graphics} 
\usepackage{graphicx}
\usepackage{epsfig} 
\usepackage{mathptmx} 
\usepackage{times} 
\usepackage{amsmath} 
\usepackage{amssymb}  
\usepackage{booktabs,tabularx}
\usepackage{multicol}
\usepackage{multirow}
\usepackage{xcolor}
\definecolor{fig_2_valid}{RGB}{0,150,0}
\definecolor{fig_2_invalid}{RGB}{255,0,0}

\usepackage{xspace}
\newcommand*{\eg}{e.g.\@\xspace}
\newcommand*{\ie}{i.e.\@\xspace}
\newcommand*{\etal}{et al.\@\xspace}
\usepackage{rotating}

\usepackage{xcolor}
\usepackage{color,soul}
\definecolor{darkgreen}{rgb}{0.0, 0.8, 0.0}

\author[1]{Amirhossein Kardoost}
\author[2,3]{Sabine M\"uller}
\author[4]{Joachim Weickert}
\author[1]{Margret Keuper}
\affil[1]{Data and Web Science Group, University of Mannheim, Mannheim, Germany}
\affil[2]{Fraunhofer ITWM, Competence Center HPC, Kaiserslautern, Germany}
\affil[3]{Fraunhofer Center Machine Learning, Germany}
\affil[4]{Mathematical Image Analysis Group, Saarland University, Saarbr\"ucken, Germany}

\begin{document}
\title{Object Segmentation Tracking\\ from Generic Video Cues}

%
%

\maketitle

\begin{abstract}

We propose a light-weight variational framework for online tracking of object segmentations in videos based on optical flow and image boundaries.
While high-end computer vision methods on this task rely on sequence specific training of dedicated CNN architectures, we show the potential of a variational model, based on generic video information from motion and color. Such cues are usually required for tasks such as robot navigation or grasp estimation. We leverage them directly for video object segmentation and thus provide accurate segmentations at potentially very low extra cost. 
Our simple method can provide competitive results compared to the costly CNN-based methods with parameter tuning.
Furthermore, we show that our approach can be combined with state-of-the-art CNN-based segmentations in order to improve over their respective results. We evaluate our method on the datasets DAVIS$_{16,17}$ and SegTrack~v2.

\end{abstract}

\IEEEpeerreviewmaketitle

\section{Introduction}
\label{sec:introduction}
Object detection and segmentation play a crucial role in applications such as grasp estimation~\cite{c35}, affordance detection~\cite{c40} or human robot interaction~\cite{c34}. While these steps are in general challenging on their own, they become even more so when we assume automotive settings in dynamic environments. Then, potentially moving objects of interest are to be segmented and tracked from video under camera ego-motion. High-end computer vision algorithms on this task usually rely on object and video specific training~\cite{c1,c2,c3} of convolutional neural networks (CNNs) and show, with few exceptions, limited applicability to online settings while they come at high computational costs.

Despite generally good results for example on the challenging DAVIS video segmentation benchmark~\cite{c4,ranet}, the boundary localization as well as occlusion handling are far from being solved by these approaches. 
However, off-the-shelf deep learning based approaches to low-level tasks such as boundary prediction~\cite{c5,c6} and optical flow estimation~\cite{c7,c8} produce highly accurate image and motion boundaries. At the same time, such low level information is a basic component in state-of-the-art approaches to robot navigation~\cite{c36,c37,c38}, grasp estimation~\cite{c35} and visual SLAM~\cite{c39}. Thus, their computation comes at little to no extra cost in many practical settings.

In this paper, we provide a light-weight variational formulation that can leverage low-level cues such as boundary estimates and optical flow estimations from generic models and incorporate them into a simple frame-by-frame label propagation framework. 

Our model facilitates the segmentation of fine details and thin structures. 
    
Furthermore, since our framework allows for a modular integration of low-level cues, it can function as an evaluation platform for such cues w.r.t. video segmentation applications. 

None of the currently evaluated optical flow or boundary estimation methods are trained or finetuned on the relevant datasets used in this paper. We thus prove the potential of \emph{generic} low-level cues for object segmentation tracking and show that the gap to highly optimized CNN methods is actually small. 
    
Additionally, we evaluate the proposed variational method as a postprocessing step for such highly optimized CNN-based models currently defining the state-of-the-art on the DAVIS$_{16}$ and DAVIS$_{17}$ dataset~\cite{c4} and show an improvement of the segmentation quality in this scenario. This experiment proves that off-the-shelf boundary and motion estimates actually carry complementary information, currently not captured in dedicated CNN-based methods. 
    
In summary, our main contributions are:
\begin{itemize}
\item[--] We provide a light-weight formulation for video object segmentation using variational label propagation. Once optical flow and boundary estimates are computed (\emph{without sequence/dataset specific finetuning}), our approach facilitates the online generation of tracked video object segmentations within milliseconds. 
\item[--] We incorporate optical flow and color features to facilitate the retrieval of lost objects due to intermediate tracking mistakes or full object occlusion.
\item[--] We study the effect of different state-of-the-art boundary~\cite{c9,c6,c5} and optical flow estimation methods in our proposed formulation~\cite{c8,c7}.
\item[--] Our approach can be used to refine state-of-the-art CNN-based results, whenever available.
\end{itemize}
	
We evaluate our method on the video object segmentation datasets DAVIS$_{16}$~\cite{c4}, DAVIS$_{17}$~\cite{c10} and SegTrack~v2~\cite{c11} and provide an ablation study on the impact of all employed cues on DAVIS$_{16}$. Our approach yields competitive results on SegTrack~v2 and can compete with several but not all CNN-based methods on the DAVIS benchmarks. Used as a postprocessing, our formulation allows to improve over existing results and provides fine object details.

\begin{figure*}[t]
		\centering
			\includegraphics[width=0.75\textwidth]{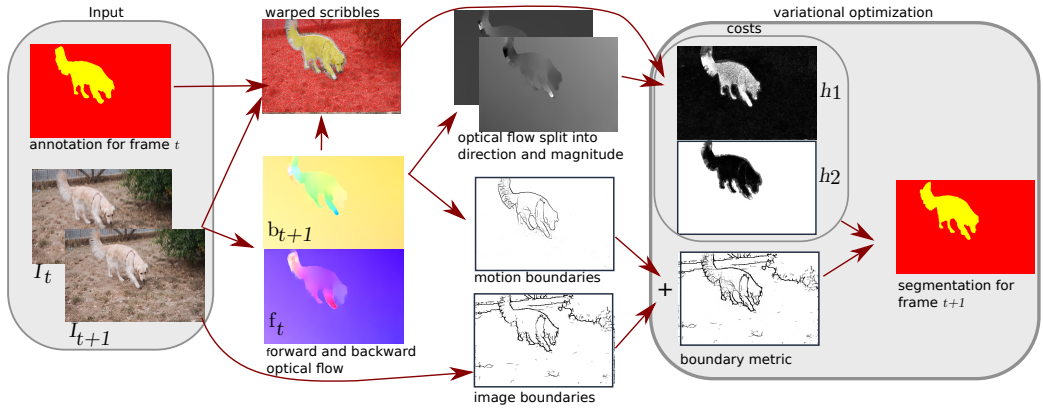}
		\caption{Visualization of the proposed workflow. Starting from the images in frame $t$ and $t+1$ and an initial annotation, scribbles are extracted based on optical flow. Then, warped scribbles, image color and optical flow values are used to generate label costs and boundary estimates to be fed into a variational segmentation framework which generates the full segmentation of frame $t+1$.
}
\label{fig:pipeline}
\end{figure*}

\subsection{Related Work}
\textcolor{black}{    
Various variational formulations have been proposed to multi-label segmentation in still images, \eg \cite{c12,c14}. In \cite{c12}, image segmentation from user scribbles is addressed in a variational framework considering the spatial and color information. In \cite{c15} such methods have been applied to produce dense video segmentations from sparse seeds in a frame-by-frame manner, based on automatically generated seeds from point trajectories \cite{c16,c17}. In contrast to \cite{c12,c15}, we directly introduce highly informative low-level cues into the variational formulation. 
Our variational formulation is derived from~\cite{c12}. Yet, we don't require user scribbles and use optical flow to propagate labels semi-densely across frames.}

Label propagation by optical flow has been previously used for example in \cite{c18,c19,c1,efficient_video}. 
Unlike \cite{c19}, which exclusively utilize temporal coherence, \cite{c1} only uses color consistency. 
Our approach employs optical flow to propagate the labels through consecutive frames and, additionally, to provide information for the data term of our formulation (\ref{var_formulate}).

The problem of label propagation in videos has also been addressed by deep learning approaches \cite{c2,c3,c20,c21,c34}. Such networks are trained on specific datasets as well as the first frame annotation of a sequence to produce segmentation of subsequent frames. Optical flow magnitudes are employed as additional input for the network \eg~in \cite{c3}, to provide additional saliency cues. However, the exact localization quality of optical flow is hardly used. Notably, \cite{c22} use optical flow to create patch correspondences in a video to improve the training of deep neural networks. Jampani~\etal~\cite{jampani2016video} use the similarity of features in neural networks to disseminate information in videos. 
In \cite{c21}, a spatio-temporal Markov Random Field model is defined over pixels to produce temporally consistent video object segmentation. In their approach, spatial dependencies among pixels are encoded by a CNN trained for the specific target sequence. In contrast, the OSVOS-S approach from Maninis~et al.~\cite{c21} can be considered to be fully complementary. They propose a one shot video object segmentation framework which explicitly does not rely on any temporal consistency within the data, such that object occlusions and disocclusions can be handled particularly well. In contrast, OSVOS-S ~\cite{c21} successively transfers generic, pretrained, semantic information to the task of video object segmentation by learning the appearance of the annotated (single) object of the test sequence. To show the benefit of our model for the refinement of CNN predictions, we particularly evaluate on OSVOS-S which can be considered most complementary.
    

\section{Proposed Approach}
In the following, we describe the details of our approach. Fig.~\ref{fig:pipeline} gives an overview of its workflow. We assume that an image sequence $I_1,\dots,I_\ell$ is given, where $\ell$ is the number of frames. 
We further assume that we are given a full annotation $\mathcal{S^\text{\it{t}}}=\cup_{i=1}^n\mathcal{S}^\text{\it{t}}_i$ 
with $n$ segments for frame $t$, where $\mathcal\cap_{i=1}^n\mathcal{S}^\text{\it{t}}_i=\emptyset$. The task is to subsequently infer the segmentation of the remaining frames using this first annotation. The proposed method computes optical flow in forward and backward direction to infer label scribbles for $I_{t+1}$ from $I_t$ and $\mathcal{S^\text{\it{t}}}$. Optical flow is also used, along with pure image information, to generate costs for all labels and to extract motion boundaries for exact object delineation. In conjunction with generic image boundaries, these cues are used to generate the full segmentation of $I_{t+1}$ using a variational formulation.

\subsection{Confident Label Propagation with Optical Flow}\label{sec:flow}

\begin{figure}[t]
		\begin{tabular}{l}
		\scriptsize
           \hspace{2.4cm}{\scriptsize  \textcolor{fig_2_valid}{$d(\mathbf{f},\mathbf{b},\mathbf{y_1})=0$}\hspace{0.3cm}\textcolor{fig_2_invalid}{$d(\mathbf{f},\mathbf{b},\mathbf{y_2})>0$}}
			\vspace{-0.05cm}\\
			\includegraphics[height=1.55cm]{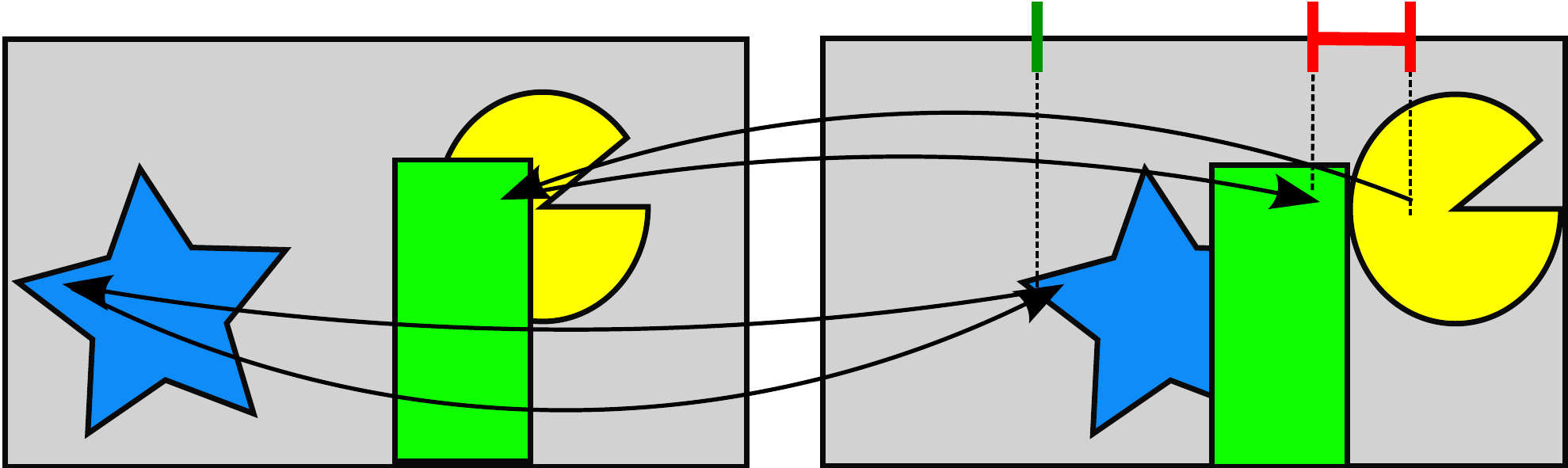}	\vspace{-0.05cm}\\
			\hspace{0.065\textwidth}$I_t$\hspace{0.125\textwidth}  $I_{t+1}$\\
			\includegraphics[height=1.45cm]{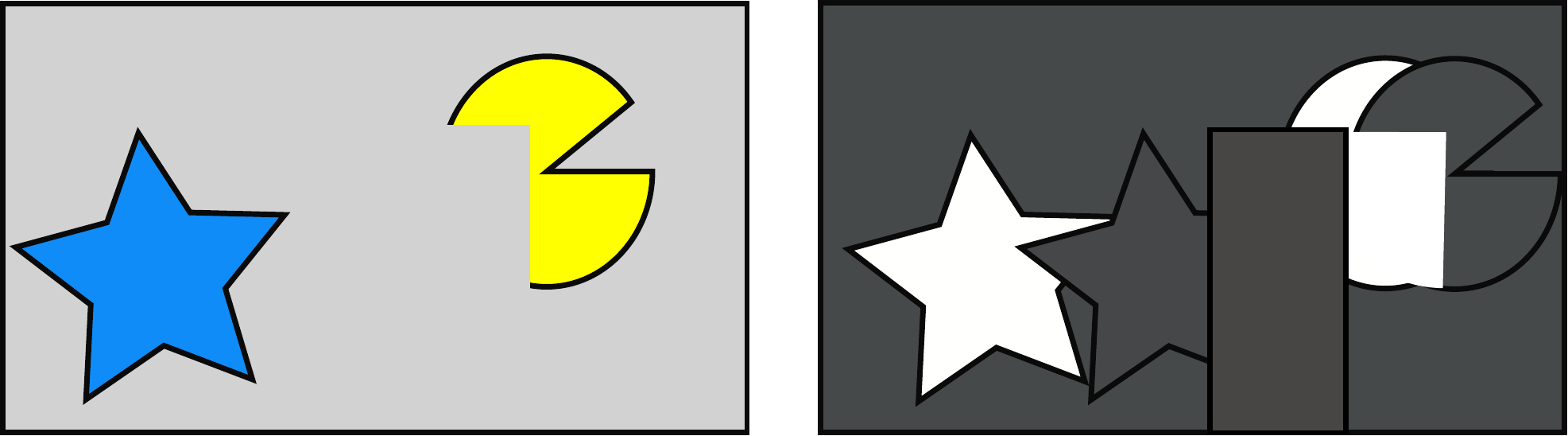}\hfill\includegraphics[height=1.45cm]{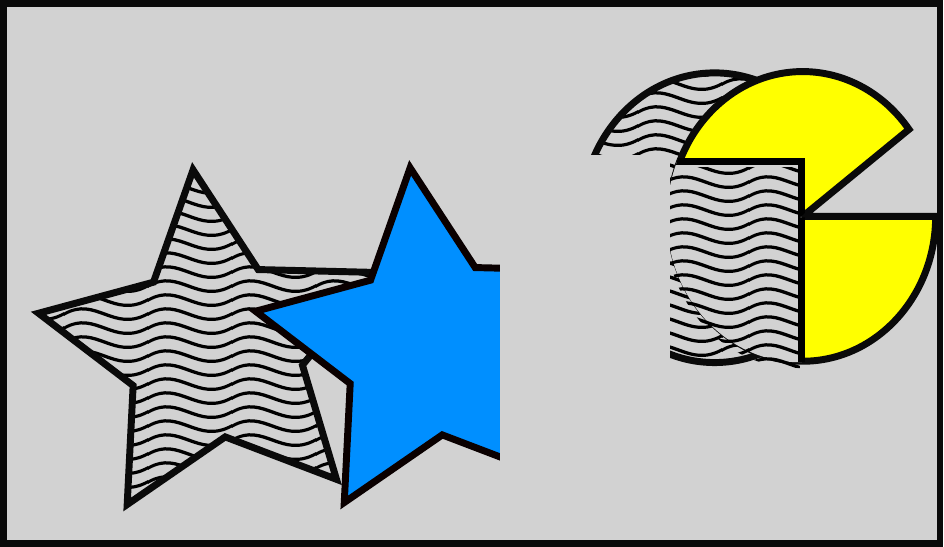}
						\vspace{-0.05cm}\\
			annotation for $I_t$\hspace{0.1cm} flow inconsistencies\hspace{0.2cm} 
			propagated labels
	\end{tabular}
	\caption{Visualization of the forward-backward consistency of the optical flow and the employed label warping. For input frames $I_t$ and $I_{t+1}$ (row 1), we check the point motion according to the backward and forward optical flow fields $\mathbf{b}_{t+1}$ and $\mathbf{f}_{t}$ for cycle consistency. For disoccluded points $\mathbf{y}$ in $I_{t+1}$ the distance $d(\mathbf{f},\mathbf{b},\mathbf{y})$ is large. In corresponding regions, no labels can be propagated.
}
    \label{fig:ofvis}
\end{figure}

The optical flow field $\mathbf{f} : \Omega\rightarrow\mathbb{R}^2$ is a function assigning a 
displacement vector to every point in the image domain $\Omega$. For every point $\mathbf{x}\in\Omega$ 
in frame $t$, the optical flow $\mathbf{f}_{t}(\mathbf{x})$ is the displacement to the most likely 
location of $\mathbf{x}$ at time ${t+1}$. Similarly, the backward optical flow $\mathbf{b}_{t+1}(\mathbf{y})$ 
of any point $\mathbf{y}\in\Omega$ in frame $t+1$ is the displacement to its most likely location 
in frame $t$. For points $\mathbf{y}$ that are visible in both frames $t$ and ${t+1}$, the distance ${d(\mathbf{f},\mathbf{b},\mathbf{y}) =
\|\mathbf{y}-\mathbf{f}_t(\mathbf{b}_{t+1}(\mathbf{y}))\|_2}$ equals to zero, see Fig.~\ref{fig:ofvis}. For those points, the label from location $\mathbf{b}_{t+1}(\mathbf{y})$ in frame $t$ can be directly transferred to location $\mathbf{y}$ in frame~$t+1$. 
Whenever a point $\mathbf{y}$ is occluded in frame
$t$, this is no longer valid. In these cases, $d(\mathbf{f},\mathbf{b},\mathbf{y})>0$.
For those locations, the label of $\mathbf{y}$ in $t+1$ needs to be inferred from other cues.
 
For real world image sequences, optical flow estimations are often not perfectly accurate such that $d(\mathbf{f},\mathbf{b},\mathbf{y})>0$ for almost all $\mathbf{y}\in\Omega$. Thus, there is need for a heuristic on the matching confidence, which is defined by $\text{conf}(\mathbf{f},\mathbf{b},\mathbf{y})$ in \eqref{eq:conf}~\cite{c23}. In practice,  we assume the optical flow matching to be confident (i.e.~$\text{conf}(\mathbf{f},\mathbf{b},\mathbf{y})=1$) if $d(\mathbf{f},\mathbf{b},\mathbf{y})$ is sufficiently small and set
	\begin{align}\label{eq:conf}
	\text{conf}(\mathbf{f},\mathbf{b},\mathbf{y}) = 
	\begin{cases}
	1, & \text{if}\  d(\mathbf{f},\mathbf{b},\mathbf{y})<\tau,\\
	0, & \text{otherwise},
	\end{cases}
	\end{align}
	with a small threshold $\tau$. 
	For confident regions in frame $t$, labels from uniformly sampled points are propagated to frame $t+1$ and considered as scribble points. Compare Fig.~\ref{fig:pipeline} (left) for a visualization.
    These propagated labels 
    are used for the data term (cost) creation of the variational formulation (Sec.~\ref{var_formulate}). 
    Additionally, the direct warping of labels in regions with high confidence renders our approach very efficient: we only need to infer labels for a small fraction of the image. 
    The intuition on this procedure is the following: Whenever we are certain that points $\textbf{y}$ in frame $t+1$ and $\textbf{x}$ in frame $t$ refer indeed to the same real world object point, the label of $\textbf{x}$ should be propagated irrespective of any other features. It is sufficient to infer labels for non-confident regions.

\subsection{Variational Formulation}
\label{var_formulate}
    
We follow \cite{c24} and formulate the multiple label segmentation problem as minimal partitioning problem. The objective is to partition the image domain $\Omega\subset\mathbb{R}^2$ into 
$\Omega_1,\dots\Omega_n\subset\mathbb{R}^2$ such as to optimize
\begin{eqnarray}\label{maincode}
	\min_{\Omega_1, \dots, \Omega_n \subset \Omega}\frac{\lambda}{2}\sum_{i=1}^{n}\textrm{Per}(\Omega_i;\Omega) + \sum_{i=1}^{n}\int_{\Omega_i}h_i(\textbf{x})\, d\textbf{x} ,\\
	\textrm{s.t.} \quad \Omega=\cup_{i=1}^n\Omega_i, \quad \forall i\neq j \quad \Omega _i\cap\Omega_j=\emptyset \nonumber. 
\end{eqnarray}	
The potential functions $h_i:\mathbb{R}\rightarrow\mathbb{R}_+$ represent the costs for each individual pixel to be assigned to label $i$, and $\textrm{Per}(\Omega_i;\Omega)$ is the perimeter of region $i$ in $\Omega$. Usually, the perimeter is measured according to an underlying image induced metric \cite{c12}. The regularization parameter $\lambda$ steers the penalization of longer boundaries. For an image $I:\Omega\rightarrow\mathbb{R}^d_+$ with $d$ channels, a common weighting function is 
\begin{equation}
	\textrm{Per}(\Omega_i;\Omega) = \int_{\Omega_i}\exp(-\gamma|\nabla I(\textbf{x})|)\,d\textbf{x} \label{bdry}
\end{equation}
where $\nabla I$ is the Jacobian of $I$, $|\nabla I|$ denotes its Frobenius norm, and $\gamma$ is a positive scalar. \\
If partial annotations $\mathcal{S}_i\subset\Omega$ of $I$ are provided for labels $i$, the potential functions $h$ can be defined as spatially varying color or feature distributions~\cite{c12}
\begin{equation}\label{hi}
	h_i(\textbf{x})=-\log\frac{1}{|\mathcal{S}_i|}\int_{\mathcal{S}_i} G_{\rho_i} \cdot G_{\sigma}\,d\textbf{x}_{S_i},
\end{equation}	
with $G_{\rho_i} = k_{\rho_i}(\textbf{x}-\textbf{x}_{S_i})$ and $G_{\sigma} = {k}_{\sigma}(I(\textbf{x}) - I(\textbf{x}_{S_i}))$. Here $|\mathcal{S}_i|$ denotes the area occupied by label $i$, ${k}_\sigma$ and $k_{\rho_i}$ are Gaussian distributions in the feature and the spatial domain, respectively. Usually, color is used as pixel features. It can be complemented for example with cues from optical flow such as its magnitude or direction. 
    
The subscripts $\rho_i$ and $\sigma$ denote the respective standard deviations. The parameter $\rho_i$ is assigned based on the Euclidean distance of the unlabeled feature points in $I$ to each of the scribble points for label $i$ 
and $\sigma$ is assigned experimentally. By $\textbf{x}_{S_i}$ and $I(\textbf{x}_{S_i})$ we represent the position ($x, y$) and feature (\eg $RGB$) information of the partial annotation $\mathcal{S}_i\subset\Omega$ in the image $I$, respectively. Equation~\eqref{hi} shows how spatial and color features of the partial annotations are used to generate costs for each label $i$ in image $I$. 

	



\begin{figure}
	\centering
	\small
	\begin{tabular}{c@{}c@{}c@{}c@{}}
 
    \includegraphics[width=0.15\textwidth]{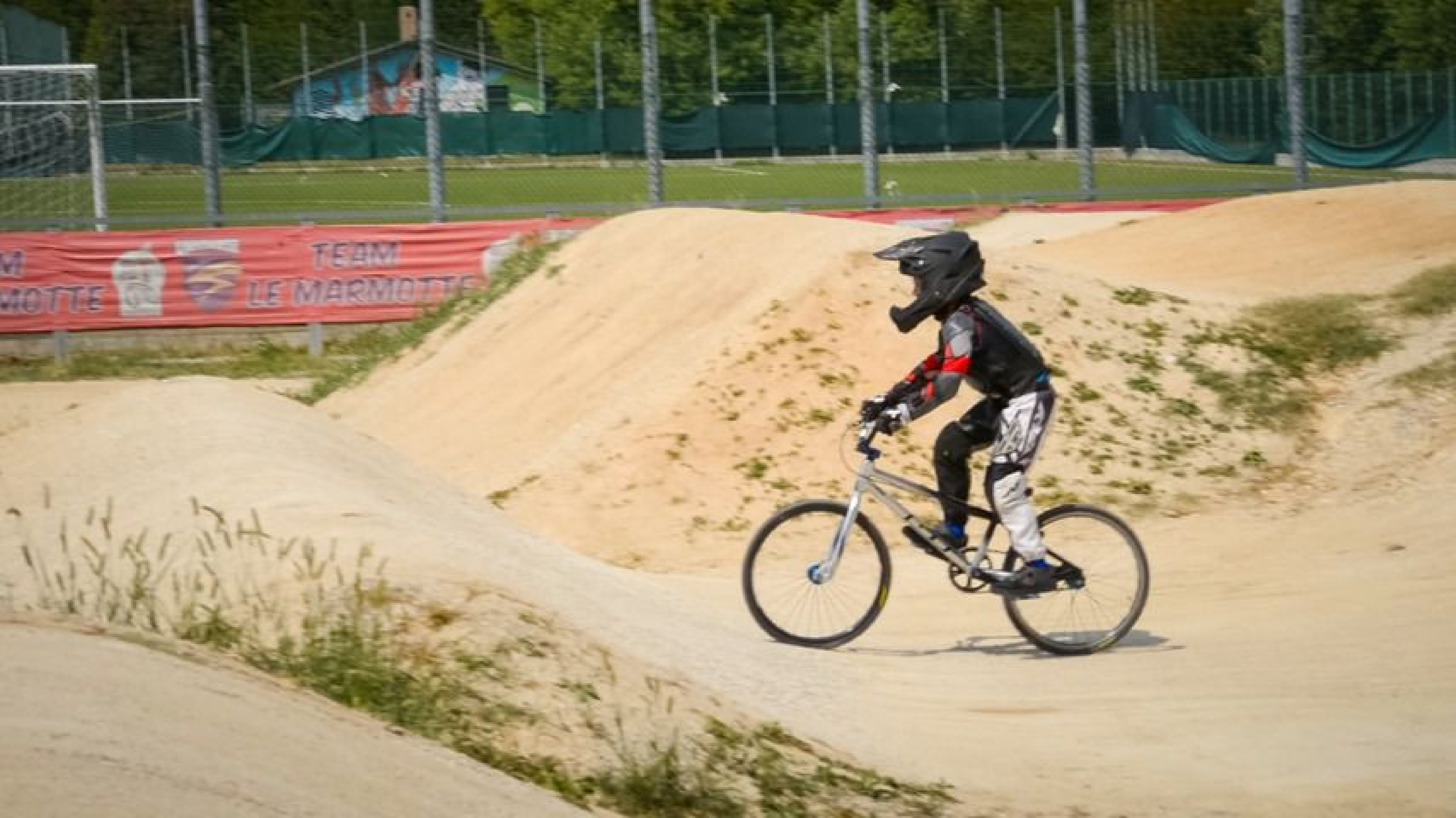}\,&
    \includegraphics[width=0.15\textwidth]{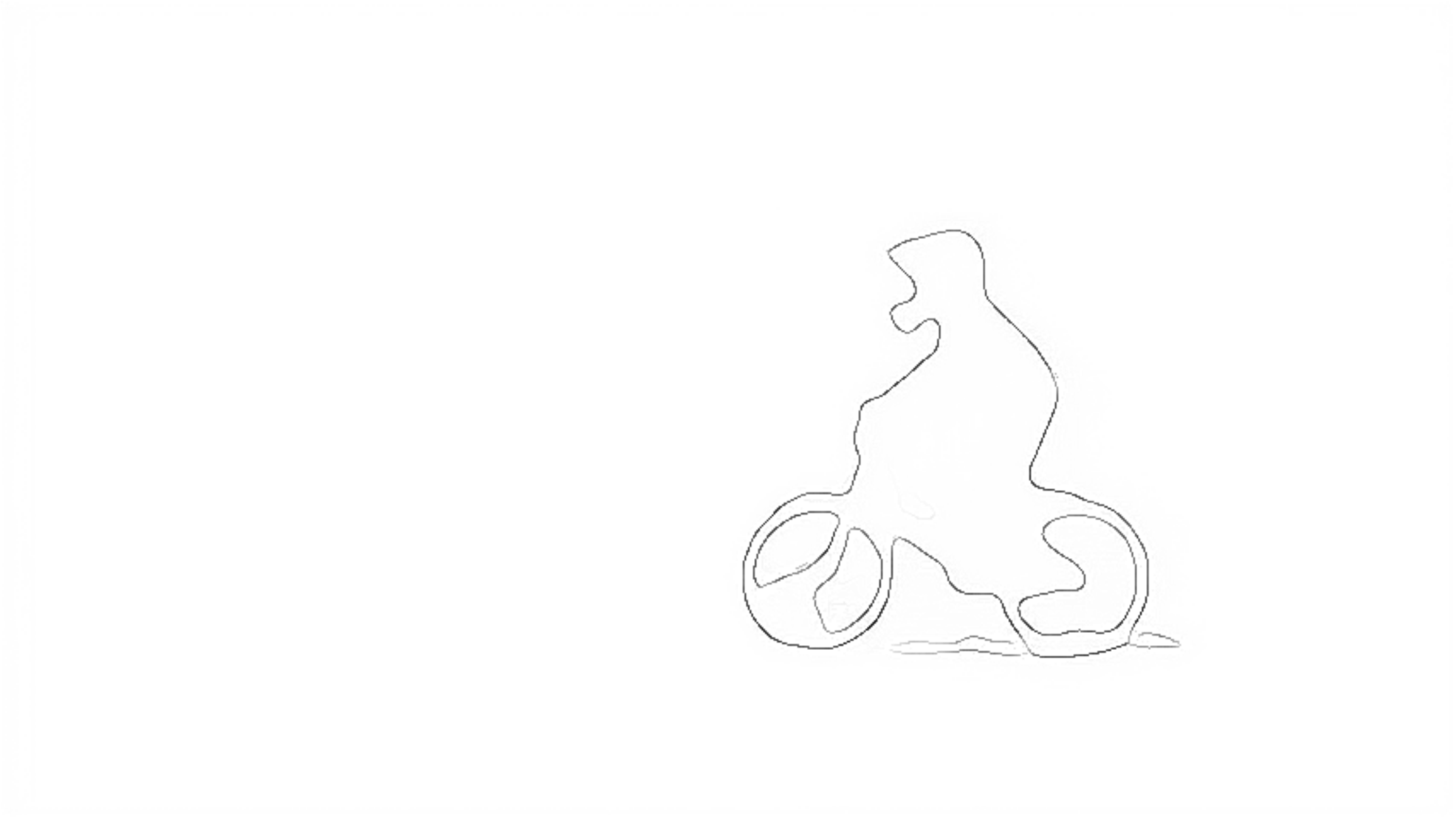}\,&
    \includegraphics[width=0.15\textwidth]{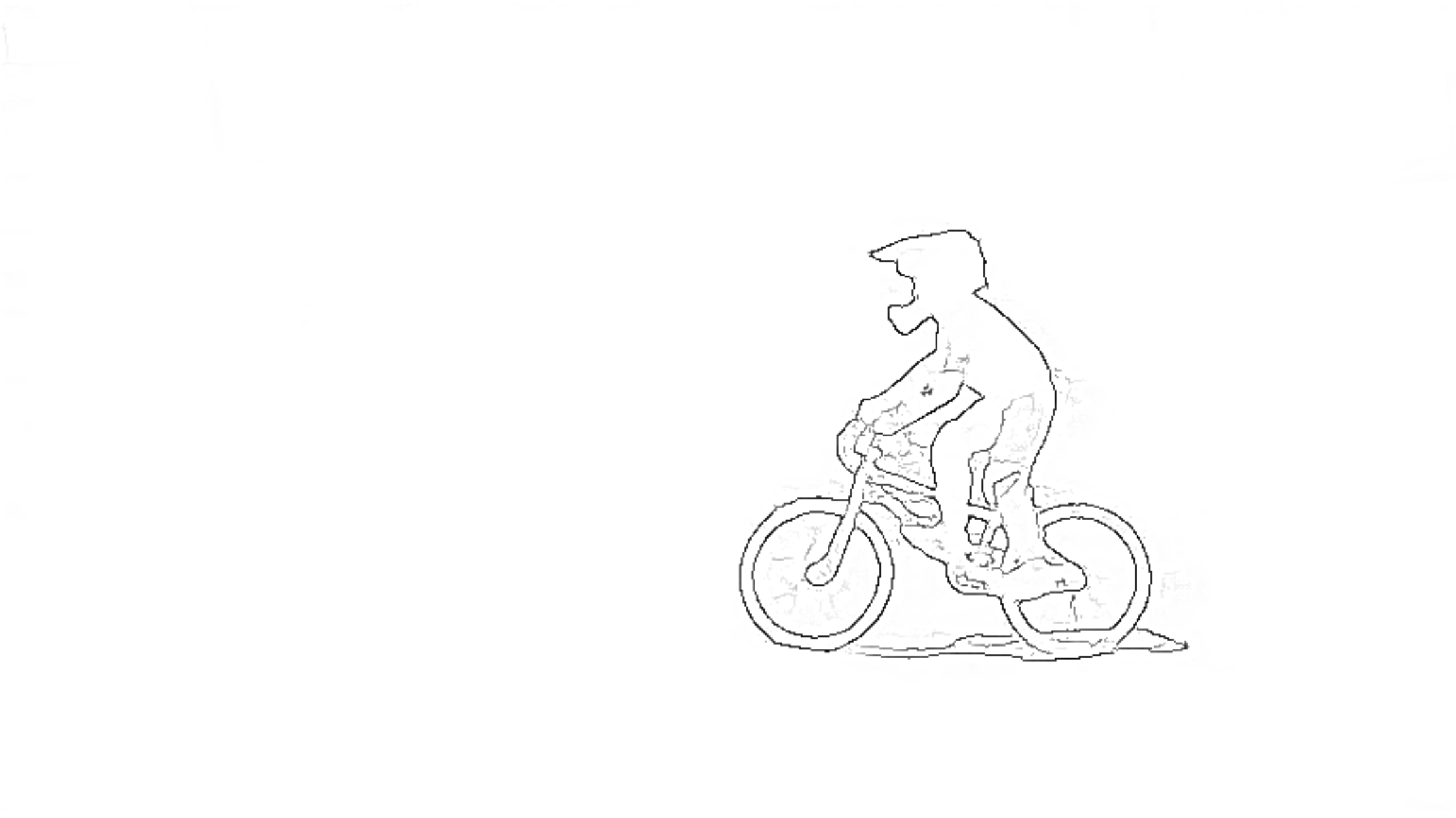}\,&\\
 input frame&FlowNet2.0~\cite{c7}&FlowNet3.0~\cite{c8}\\
    \includegraphics[width=0.15\textwidth]{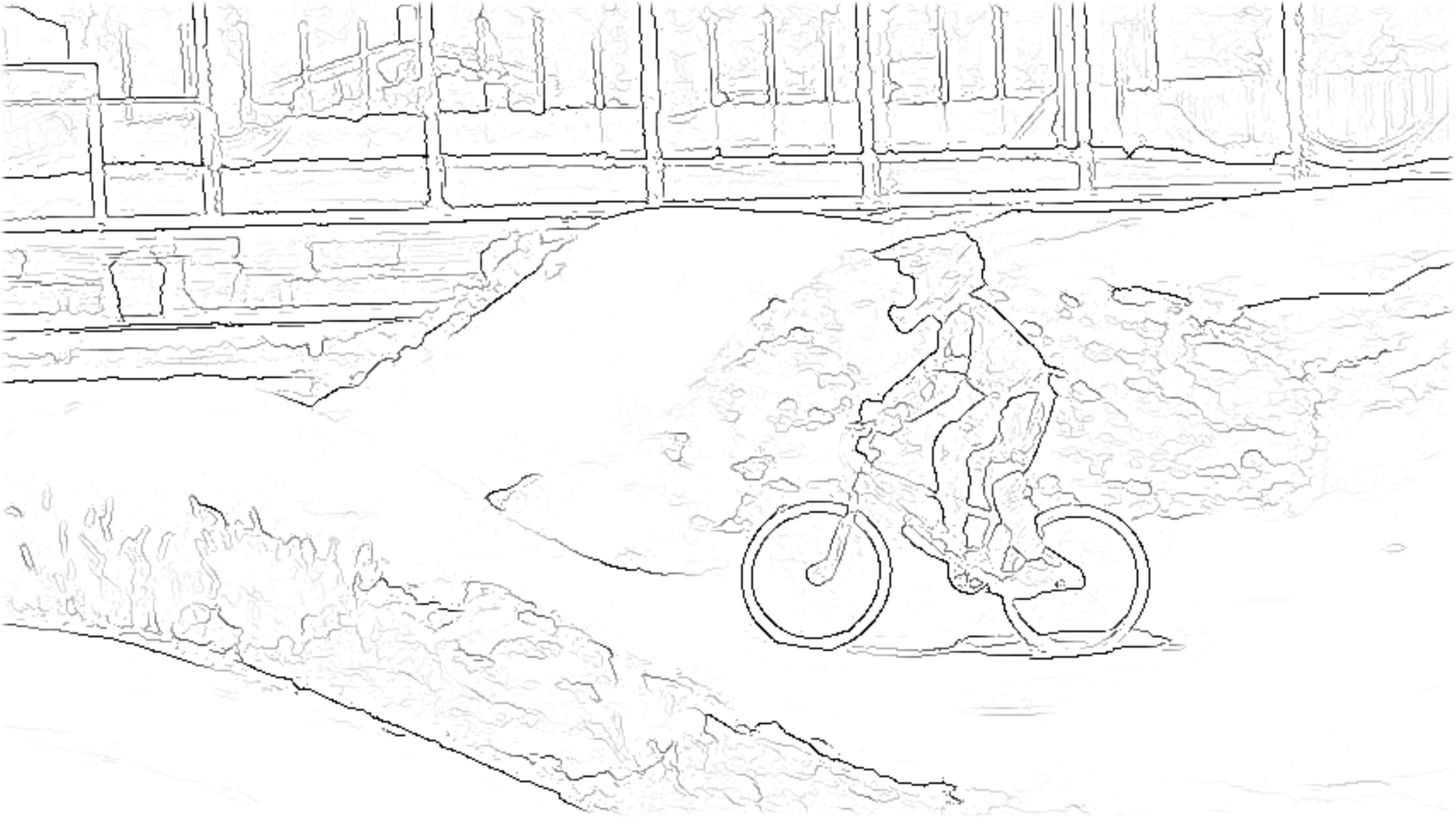}\,&
    \includegraphics[width=0.15\textwidth]{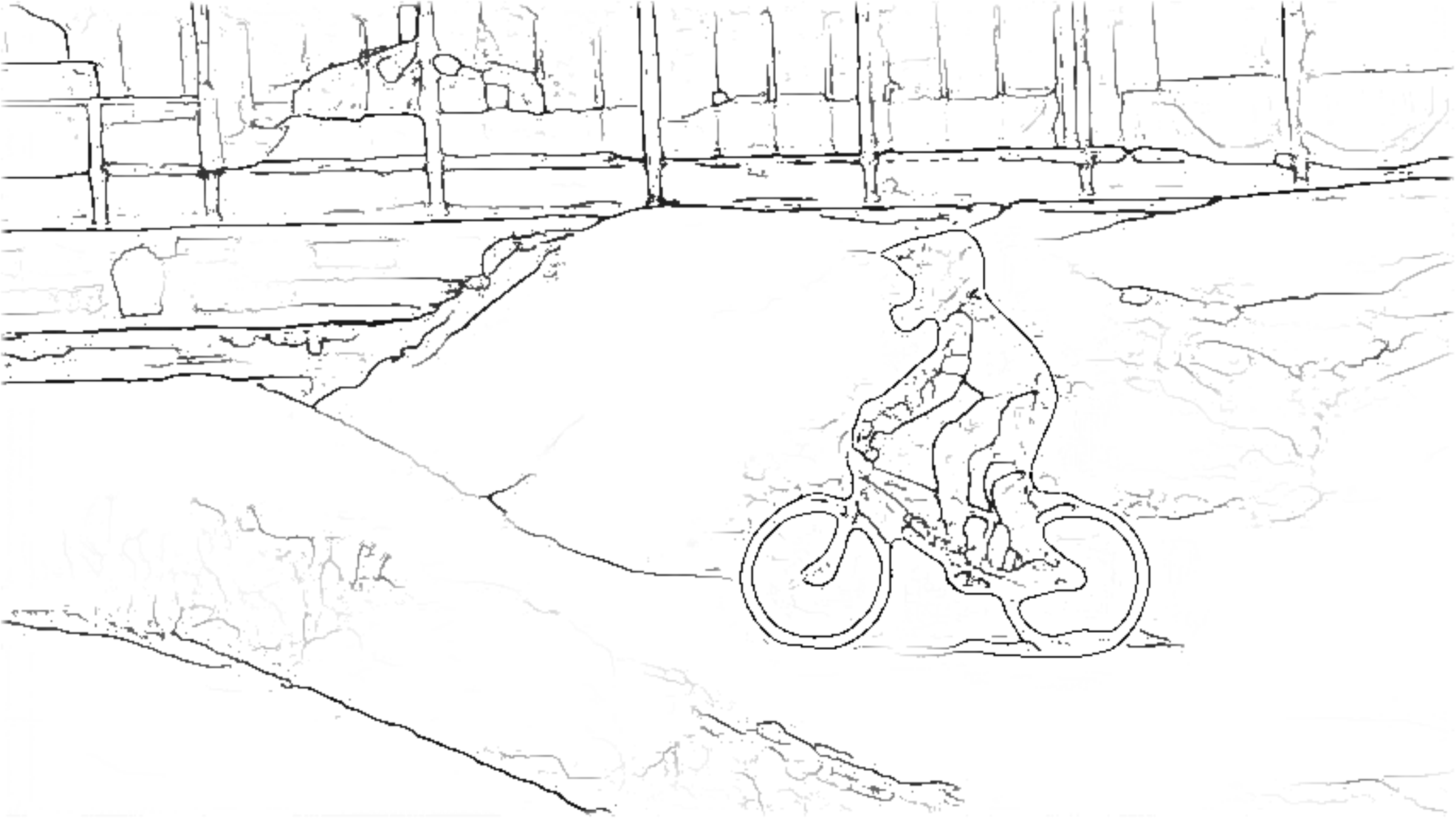}\,&
    \includegraphics[width=0.15\textwidth]{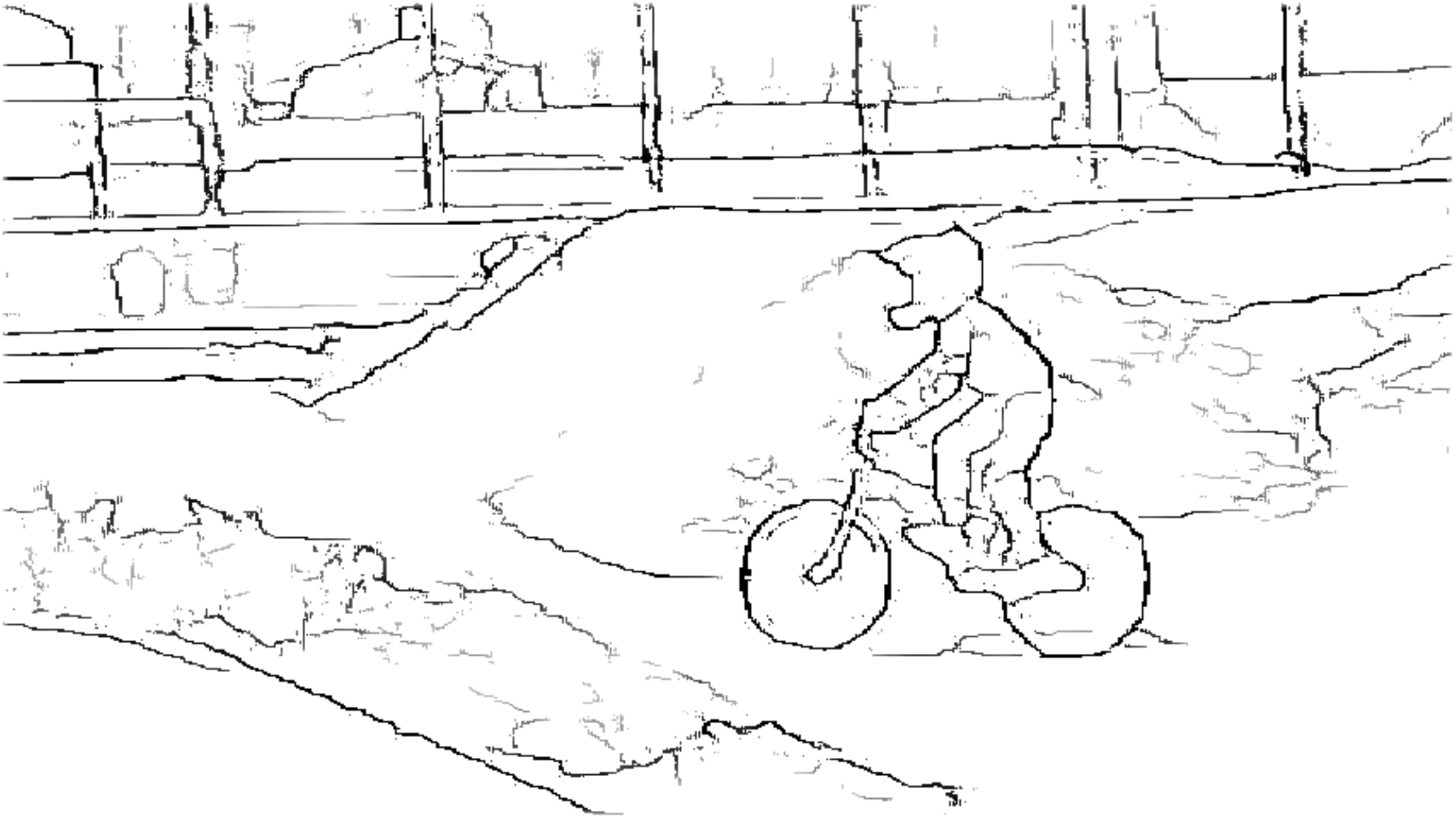}\,&\\
    $SED$~\cite{c9}&$HED$~\cite{c6}&$COB$~\cite{c5}&
 
	\end{tabular}
	\caption{We depict different boundary estimations in the ``bmx-bumps'' sequence of DAVIS$_{16}$. While FlowNet3.0~\cite{c8} directly estimates motion boundaries, we compute them from gradient magnitudes in the optical flow from FlowNet2.0~\cite{c7}.
	}
	\label{fig:motion_boundary_bmx_bumps}
\end{figure}
    
\subsection{Flow Magnitude and Flow Direction} 
\label{flowmag_flowdir}
Besides being useful for the tracking of ego-motion and 3D scene reconstruction for example by visual SLAM~\cite{c39}, optical flow information is a straight-forward cue for video label propagation. Here, we additionally leverage optical flow information in the data term $h_i$ \eqref{hi} for the creation of the label costs. Such information is expected to provide: 1) cues for object saliency, and 2) cues for the object label, since motion only changes gradually over time. 
Hence, we concatenate the normalized flow magnitude ($\mathbf{f}_{\text{mag}}$) and direction ($\mathbf{f}_{\text{dir}}$) to the original color information of the image in frame $t+1$ for the segmentation. With this additional information, 
$I:\Omega\rightarrow\mathbb{R}^3_+$ in \eqref{hi} is replaced by 
\begin{equation} \label{eq:J}
	J := \begin{bmatrix} I \\ \alpha \cdot \mathbf{f}_{\text{mag}} \\ \theta \cdot \mathbf{f}_{\text{dir}} \end{bmatrix},
\end{equation}
where $\mathbf{f}_{\text{mag}}:\Omega\rightarrow\mathbb{R_+}$, $\mathbf{f}_{\text{dir}}:\Omega\rightarrow\mathbb{R_+}$ and $\alpha$ and $\theta$ are weighting factors which are assigned as $0.5$ to account for the strong expected correlation between color values. Thus, the range of values in the $RGB$ channels are between $0$ and $255$ while values in $\mathbf{f}_\text{mag}$ and $\mathbf{f}_\text{dir}$ range between $0$ and $127.5$. Finetuning of these parameters for specific datasets is possible and will most likely improve the results. However, the proposed approach attempts not to fit such parameters to any specific dataset for simplicity.
It is important to notice that the produced optical flow estimations are not always accurate. More specifically, the flow information is produced with models trained on common optical flow benchmark datasets, which are different from the datasets we evaluate and use in our approach. Yet, our model benefits even from noisy optical flow estimations.
	
\subsection{Boundary Term} 
\label{learned_boundary_term} 
In the variational formulation from~\eqref{maincode}, the perimeter $\mathrm{Per}$ is computed based on an image induced metric such as given in~\eqref{bdry_1}. This metric can be replaced by more evolved, learning-based boundary estimations $\mathcal{E}:\Omega\rightarrow \mathbb{R}_+$ such as \cite{c9,c5,c6}. For example, \cite{c15} propose to weight the region boundaries according to pseudo-probabilities
\begin{equation}
	g(\textbf{x})=\exp(\mathcal{E}(\textbf{x})^\beta/\bar{\mathcal{E}}),
	\label{bdry_1}
\end{equation}
with $\displaystyle\bar{\mathcal{E}}:=\frac{2}{\Omega}\int_\Omega |\mathcal{E}(\textbf{x})|d\textbf{x}$ and $\beta>0$, and employ boundary estimates from \cite{c9} for the generation of object segmentations.
       
However, any approach to image (\eg HED \cite{c6}), object (\eg COB \cite{c5}), or motion (\eg FlowNet3.0~\cite{c8}) boundary estimation could be used. 
We study the effect of each of the mentioned boundary 
estimation methods on the validation set of DAVIS$_{16}$ in Sec.~\ref{par::bdry}. For this study, off-the-shelf trained models of \cite{c9,c6,c5,c8} are employed to generate boundary estimates (compare Fig.~\ref{fig:motion_boundary_bmx_bumps}).  Please notice that none of the employed models is neither trained nor finetuned on the datasets under consideration.

\begin{figure}[!t]
        \centering
	    \begin{tabular}{ @{\hspace{0.0cm}} c @{\hspace{0.1cm}}c @{\hspace{0.1cm}}c @{\hspace{0.0cm}} }
			        \includegraphics[width=0.25\columnwidth]{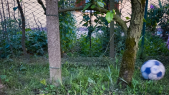}&
			        \includegraphics[width=0.25\columnwidth]{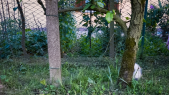}&
			        \includegraphics[width=0.25\columnwidth]{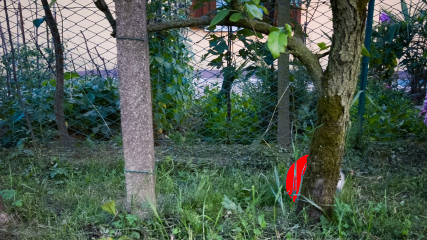}\\
			        frame 1&frame 5&frame 6
			    \end{tabular}
		    \caption{The ``soccerball'' sequence from DAVIS$_{16}$~\cite{c4} provides an example of an object reappearing after occlusion. For such objects, no labels can be propagated.}
            \label{fig:lostObject}
    \end{figure}
	
\section{Lost Object Retrieval
\label{lost_obj_ret}}
In the considered video object segmentation scenario, objects can become partially or fully occluded for several video frames. In this case, the respective label gets lost. 
Fig.~\ref{fig:lostObject} illustrates this problem: The foreground object (a soccerball) moves to the left and is partially covered by a tree. 
As it reappears, no labels can be propagated. We propose a simple approach to lost object retrieval~(LOR), which fixes this issue in many practical scenarios.
	
We create partial annotations of the missing object using the confidence values from \eqref{eq:conf} and the color information given in the annotated key frame. 
As soon as the object reappears (\ie is disoccluded) in frame $I_{t+1}$ to be segmented, the confidence of the label propagation in the respective image area should be low, since $d(\mathbf{f},\mathbf{b},\textbf{y})$ is high in case of disocclusion (see Fig.~\ref{fig:ofvis}).
	
Thus, we select all positions with low confidence for the label 
propagation as candidates for LOR. Then, we compute the color similarity of the positions in $I_{t+1}$ with the lost object's mean color extracted from the annotated key frame. Finally, we create partial labels for the object using the calculated color similarities by selecting the points with a color distance below a predefined threshold (here set to $5.0$). Such a low value is necessary to prevent wrong partial label generation and ensures that we retrieve lost objects whenever the color similarity of respective regions is high. 
This approach works well in practice. However, it might fail when different objects on a video are similar in color. Here, 
we apply LOR in the binary segmentation scenario only. 

\section{Implementation Details} 
\label{imp_det}
    
{\it Parameter Settings} Our approach has several parameters. 
The $\gamma$ in \eqref{bdry} is trivially set to $\frac{1}{255}$ and the normalization factors $\alpha$ and $\theta$ in~\eqref{eq:J} are set to $0.5$ for all datasets. The $\lambda$ in~\eqref{maincode} is determined via grid search in the interval $\{5, 10,...,60 \}$ for the first two images of each sequence: 
We assume that the object size does not change drastically in two consecutive frames, and thus select the $\lambda$ yielding the smallest deviation in size between the foreground objects in the generated segmentation and the first frame annotation.
We set $\tau$ in~\eqref{eq:conf} to a fixed value of $5$ for DAVIS$_{16}$ and DAVIS$_{17}$. For SegTrack~v2, $\tau$ is set to the mean optical flow magnitude to account for strong motion variations. 
The value $\sigma$ in~\eqref{hi} is set to $64$ for all sequences in all datasets. We expect that parameter finetuning would improve the results further. Yet, we skip this step for simplicity.
    
    
{\it Optimization}
Assigning labels to each object is an optimization problem that we solve using the iterative primal-dual algorithm of \cite{c24}. 
Stopping criteria for this iterative approach are based on a maximum number of iterations which initially is set to $3000$. It is increased to $6000$ whenever the calculated objective value in iteration $3000$ is above $600000$. The optimization stops earlier when the decrease in objective value between consecutive iterations is below $10$. The computation time for the optimization is proportional to the number of iterations and objects to be segmented.

\section{Experiments and Results}
We evaluate our method on binary and multi-object segmentation tasks on the state-of-the-art datasets DAVIS$_{16}$~\cite{c4}, DAVIS$_{17}$~\cite{c10} and SegTrack~v2~\cite{c11}.
	
\subsubsection{DAVIS Benchmark}
The original version of the DAVIS$_{16}$ benchmark \cite{c4} is focused on binary video object segmentation and consists of 50 sequences with pixel-accurate object masks. It contains different challenges such as light changes, occlusions and fast motion.
The more recent dataset DAVIS$_{17}$~\cite{c10} also includes the segmentation of multiple objects. It consists of 90 sequences and more complicated scenarios for example due to object interactions. 
The DAVIS datasets are evaluated in terms of boundary accuracy (F-measure) and Jaccard's index (i.e. intersection over union (IoU)).
    
\subsubsection{SegTrack v2 Dataset}
The SegTrack v2 dataset~\cite{c11} consists of 14 sequences with ground truth annotation per frame and object. The sequences contain different object characteristics, motion blur, occlusions and complex deformations, low resolution and quality, for both binary and multi-object scenarios. The standard evaluation metric is IoU.
    
\subsection{Ablation Study}
In the following, we evaluate the impact of the employed cues such as boundary terms, optical flow and lost object retrieval to our model on the DAVIS$_{16}$ dataset. 
    	
\subsubsection{Boundary Terms}
\label{par::bdry}
    \begin{table}[t]
		\centering
		\caption{Our results for different boundary estimation methods on the $\text{DAVIS}_{16}$ validation set. 
		Motion boundaries (MB)s from \cite{c8} are studied when combined (\textbf{w/} MB) or not combined (\textbf{w/o} MB) to each of the boundary detectors.
		}
		\label{smoothnessTerm_2016}
		\begin{tabular}{l |@{\hspace{0.4cm}} c @{\hspace{0.4cm}} c @{\hspace{0.4cm}} c @{\hspace{0.4cm}}   c}
			\toprule
			&\multicolumn{2}{c}{$\it{F(\%)}$}	& \multicolumn{2}{c}{$\it{J(\%)}$} \\ 
			\midrule
			Method	& \textbf{w/o} MB &  \textbf{w/} MB	 & \textbf{w/o} MB & \textbf{w/} MB \\ \midrule
			${SED}$ ~\cite{c9}	& $56.72$ & $64.71$	& $62.64$ & $69.09$  \\
			${HED}$	~\cite{c6}& $59.77$ & $67.33$   & $63.65$ & $70.99$  \\
		    ${COB}$	~\cite{c5}& $\mathbf{60.47}$ & 
		    $\mathbf{68.39}$ & $\mathbf{63.71}$ & 
		    $\mathbf{71.58}$  \\ \midrule
		\end{tabular}
	\end{table}
	
The boundary term used in~\eqref{bdry_1} is crucial to our approach. We evaluate segmentation results when \cite{c9} (${SED}$), \cite{c6} (${HED}$) and \cite{c5} (${COB}$) are used directly and when they are combined with motion boundaries extracted from FlowNet3.0~\cite{c8} (see Fig.~\ref{fig:motion_boundary_bmx_bumps}). In this case, motion boundaries are simply summed up before non-maximum suppression. 
All boundary estimations are generated based on existing models, including \cite{c8}, who directly estimate motion boundaries along with the optical flow. We emphasize that none of these models is trained on DAVIS$_{16,17}$ nor SegTrack v2.
In Tab.~\ref{smoothnessTerm_2016}, we report the resulting F-measure ($\it{F}$) and Jaccard's index ($\it{J}$) values. The combination of $COB$ and motion boundaries works best. All further results are based on this setting. 

\subsubsection{Flow Estimation Methods} 
Optical flow information is a central component of our model. It is used to 1) generate scribble points for subsequent frames, 2) compute the data terms $h_i$ in the variational optimization, and 3) to compute motion boundaries to complement generic image boundaries (compare Fig.~\ref{fig:motion_boundary_bmx_bumps}).
Only few optical flow methods produce motion boundary estimates directly as an additional output, which is why our setup is based on FlowNet3.0~\cite{c8}. However, motion boundaries can be computed from strong gradients of any estimated optical flow field, such as generated by~\cite{c25,c26,c7}. Thus, we compare here the performance of our full model when we replace all optical flow information from  FlowNet3.0~\cite{c8} by FlowNet2.0~\cite{c7}. The results in Tab.~\ref{dataterm_2016} show a decrease in the segmentation accuracy. Since the difference in optical flow quality itself is known to be small between the two approaches, the decrease in segmentation quality indicates a rather strong impact of the better motion boundaries from FlowNet3.0. 

 \begin{table}
		\centering
		\caption{Our results for train and validation sets of DAVIS$_{16}$ when: 1. using FlowNet2.0~\cite{c7} instead of FlowNet3.0~\cite{c8}, 2. not using lost object retrieval (\textbf{w/o} LOR), 3. employing different components of spatial, color and optical flow information.
		} 
		\label{dataterm_2016}
		\begin{tabular}{l| c    c   c   c}
			\toprule
			&\multicolumn{2}{c}{$\it{F(\%)}$}	& \multicolumn{2}{c}{$\it{J(\%)}$}\\ 
			\midrule
			Method & train & val & train & val \\ 
			\midrule
FlowNet2.0 ~\cite{c7} & $71.98$ & $68.05$  & $75.64$	& $71.46$ \\
			\midrule
			\textbf{w/o} LOR	& 67.50	 	   & 63.81	          & 72.21 	     & 68.30           \\
			\midrule
			\textbf{w/o} $\mathbf{f}_{\text{mag}}$ + $\mathbf{f}_{\text{dir}}$& $69.66$ 	& $58.17$	 & $68.36$ 	   & $59.51$\\
			\textbf{w/o} $\mathbf{f}_{\text{dir}}$ & $71.93$ 	& $67.19$ & $75.24$ & $69.45$  \\
			\emph{our full model}& $\mathbf{73.49}$& $\mathbf{68.39}$      & $\mathbf{77.68}$  & $\mathbf{71.58}$  \\ 
			\bottomrule
		\end{tabular}
	\end{table}

\subsubsection{Lost Object Retrieval} 
In Tab.~\ref{dataterm_2016}, we evaluate the impact of lost object retrieval (LOR) to our method. The numbers indicates a significant improvement of the segmentations due to LOR. 
Specifically, the results of our model improve by ${3-5}$\% on the train and validation sets of DAVIS$_{16}$ when LOR is added. 
    
\subsubsection{Data Term}
In Tab.~\ref{dataterm_2016}, we provide an ablation study on the data term creation. To do so, we remove from our \emph{full model}, the optical flow direction ({\bf w/o} $\mathbf{f}_{\text{dir}}$) and both the optical flow direction and magnitude ({\bf w/o} $\mathbf{f}_{\text{mag}} + \mathbf{f}_{\text{dir}}$). In this case, only color information is used. Our full model performs better than the two alternatives, thus both  $\mathbf{f}_{\text{mag}}$ and $\mathbf{f}_{\text{dir}}$ provide meaningful segmentation cues.

\begin{figure*}[t]
		\centering 
		\begin{tabular}{@{\hspace{0.05cm}}c@{\hspace{0.05cm}}c@{\hspace{0.05cm}}c@{\hspace{0.05cm}}c@{\hspace{0.05cm}}c@{\hspace{0.05cm}}c@{\hspace{0.05cm}}c@{\hspace{0.05cm}}c@{\hspace{0.05cm}}}

            \includegraphics[width=0.12\textwidth]{cows_00046.pdf}& 
            \includegraphics[width=0.12\textwidth]{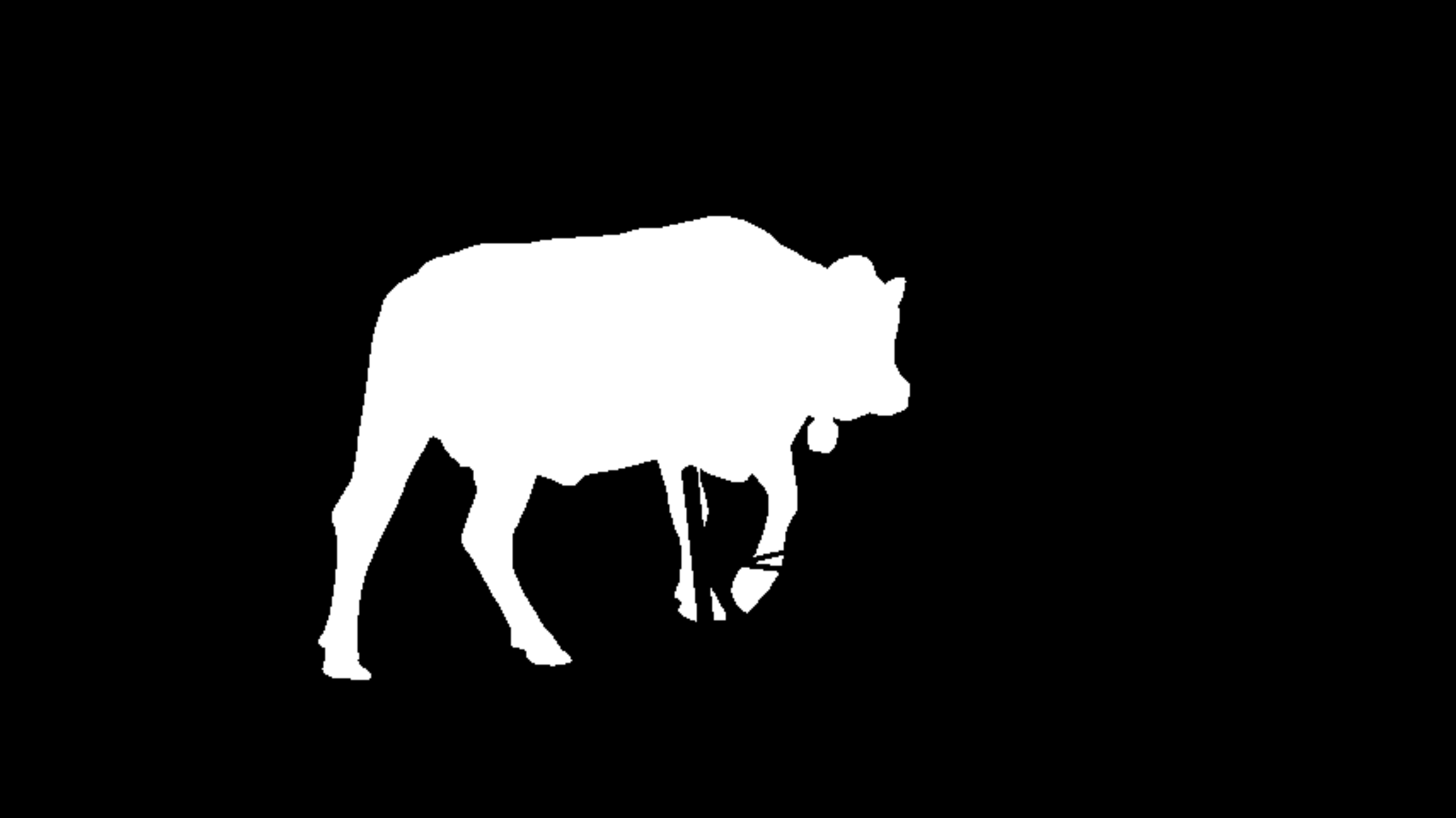}& 
            \includegraphics[width=0.12\textwidth]{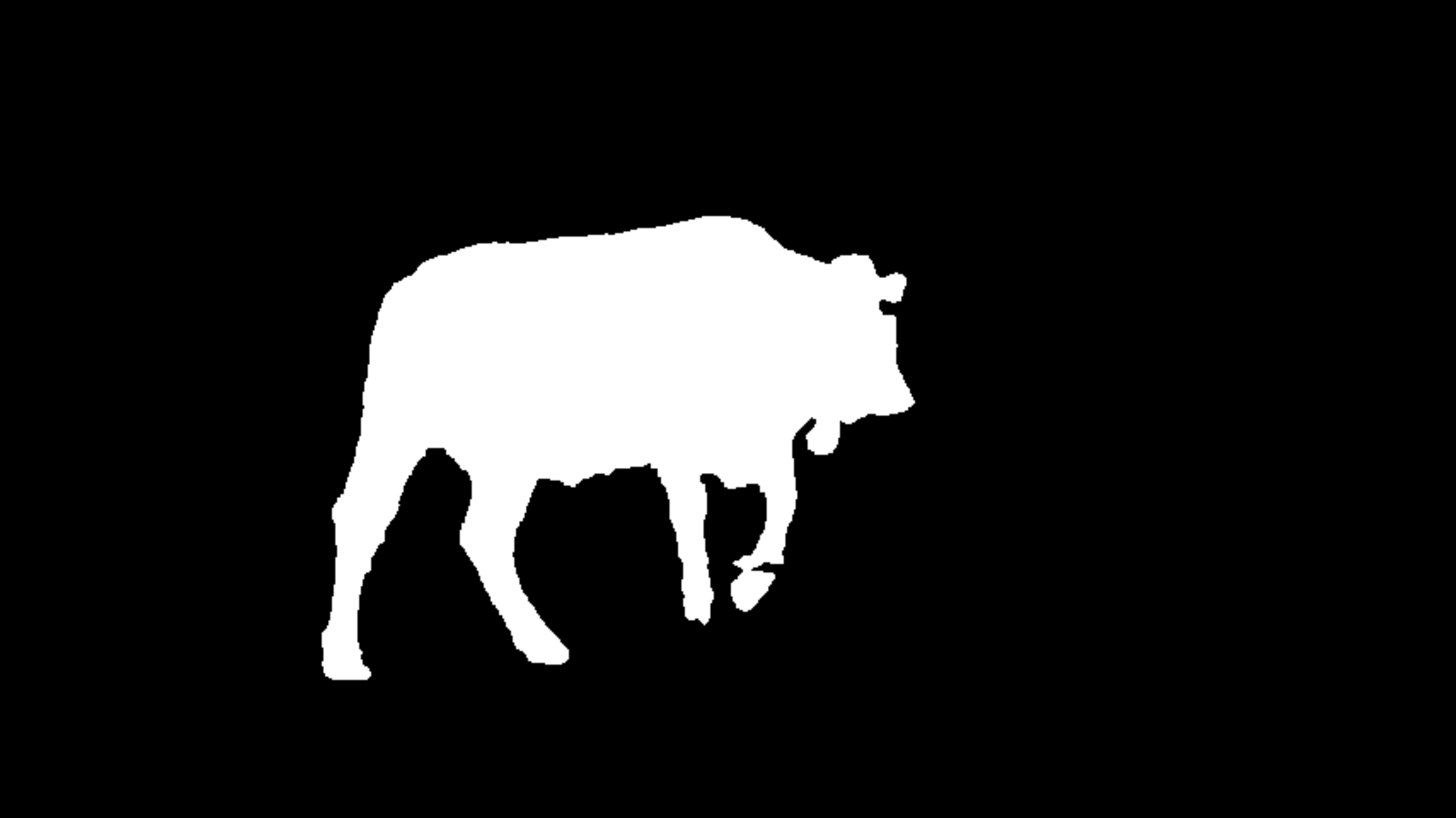}&
            \includegraphics[width=0.12\textwidth]{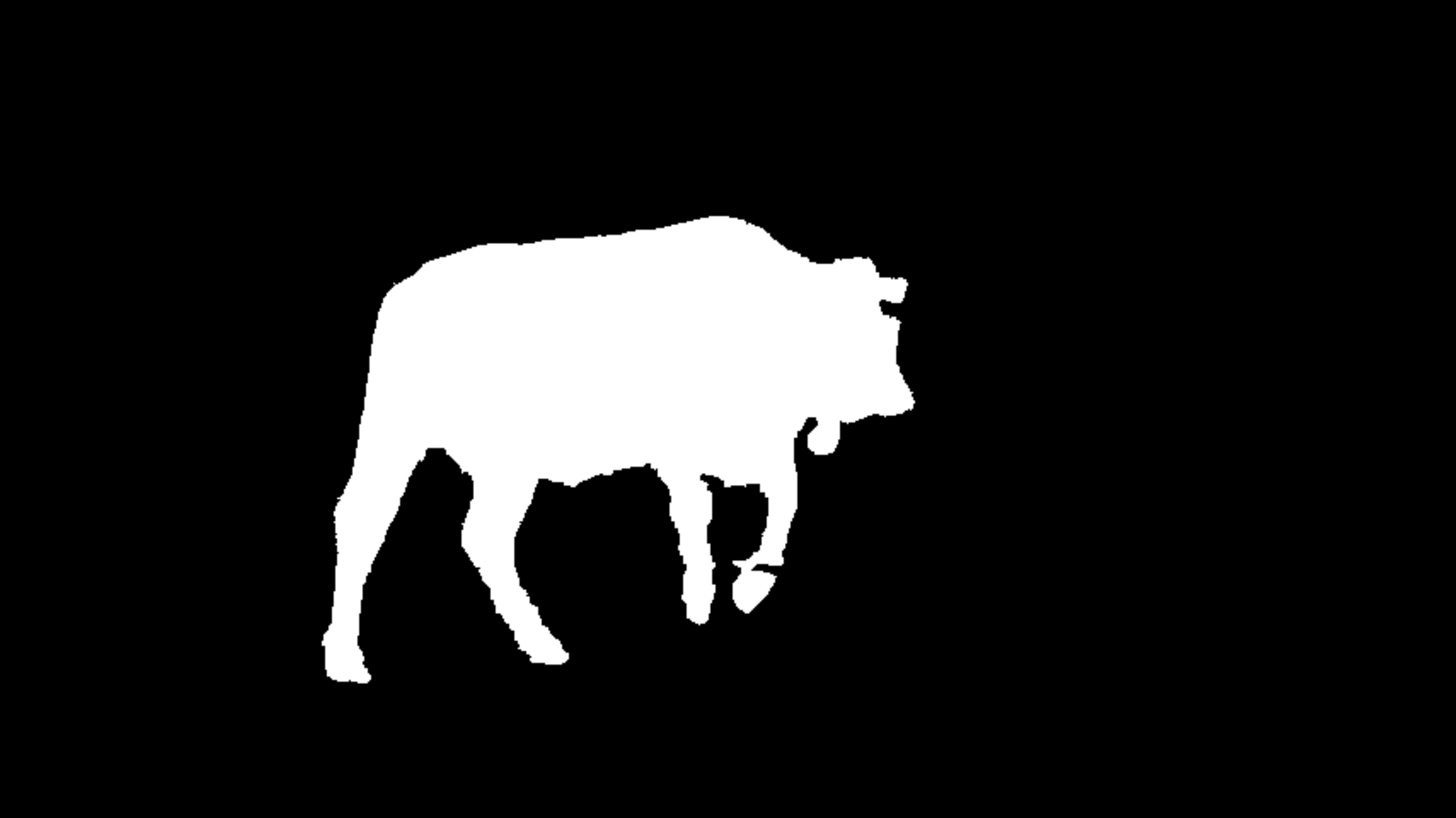}&
            \includegraphics[width=0.12\textwidth]{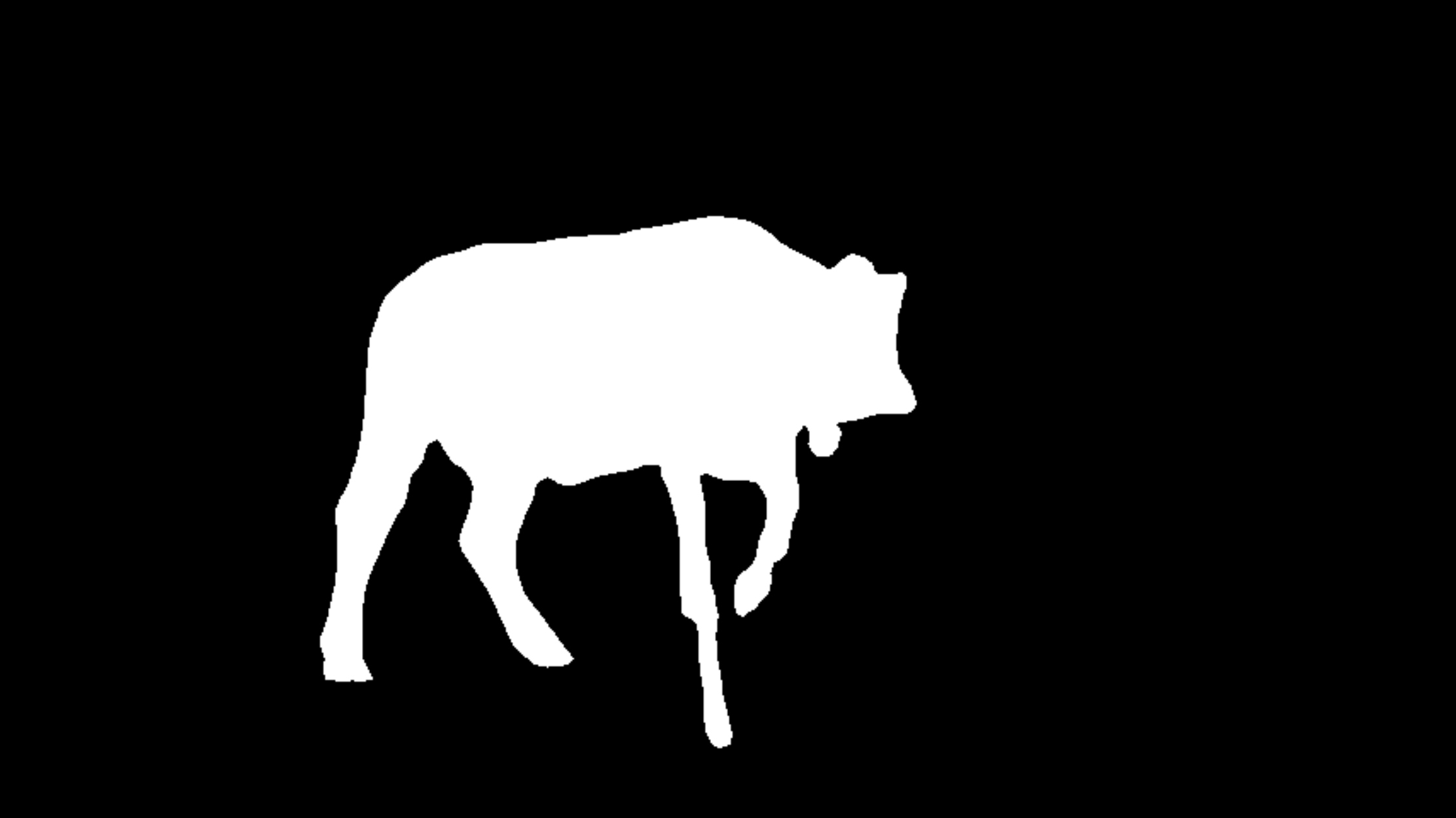}& 
            \includegraphics[width=0.12\textwidth]{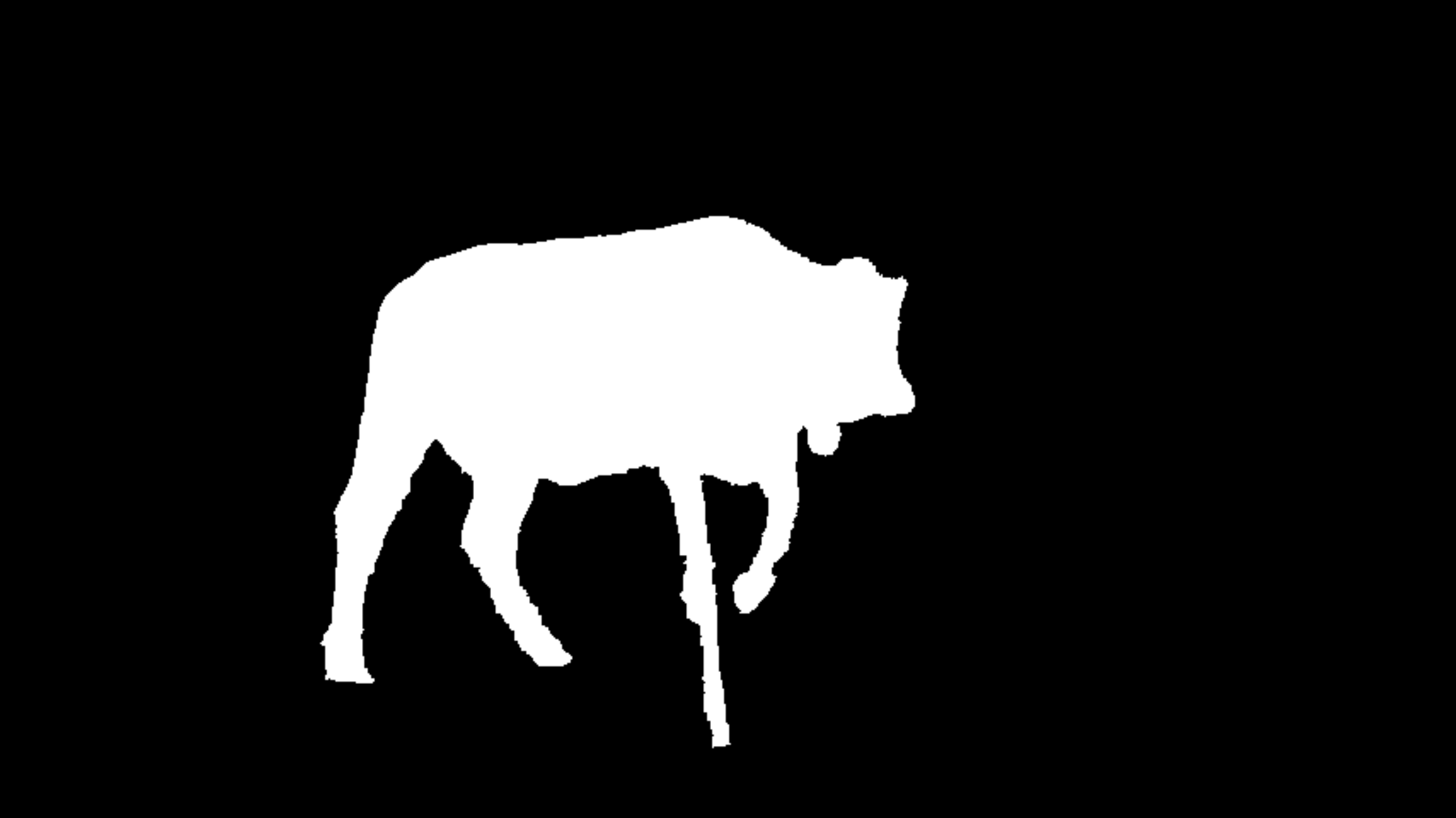}&
            \includegraphics[width=0.12\textwidth]{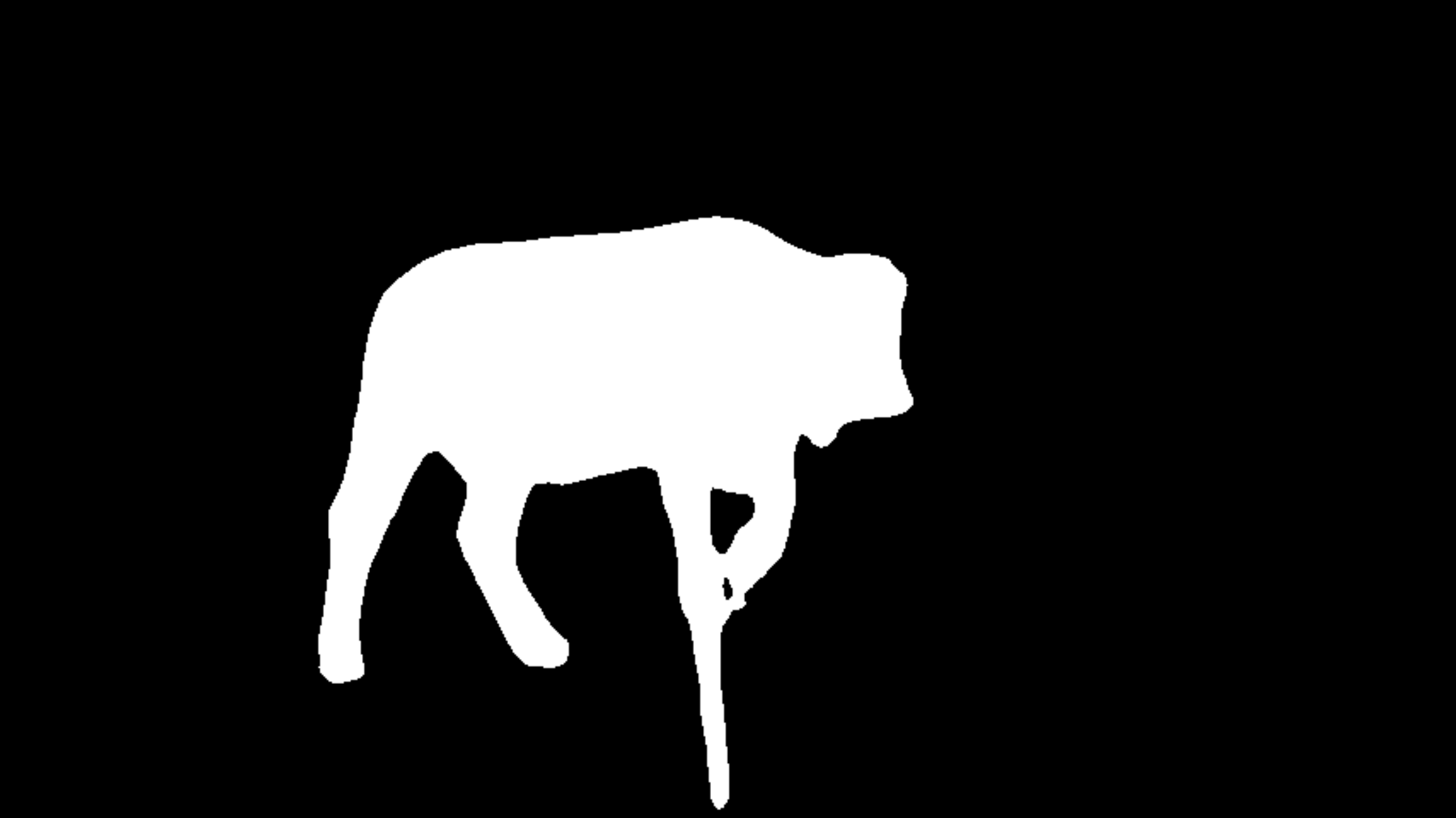}&
            \includegraphics[width=0.12\textwidth]{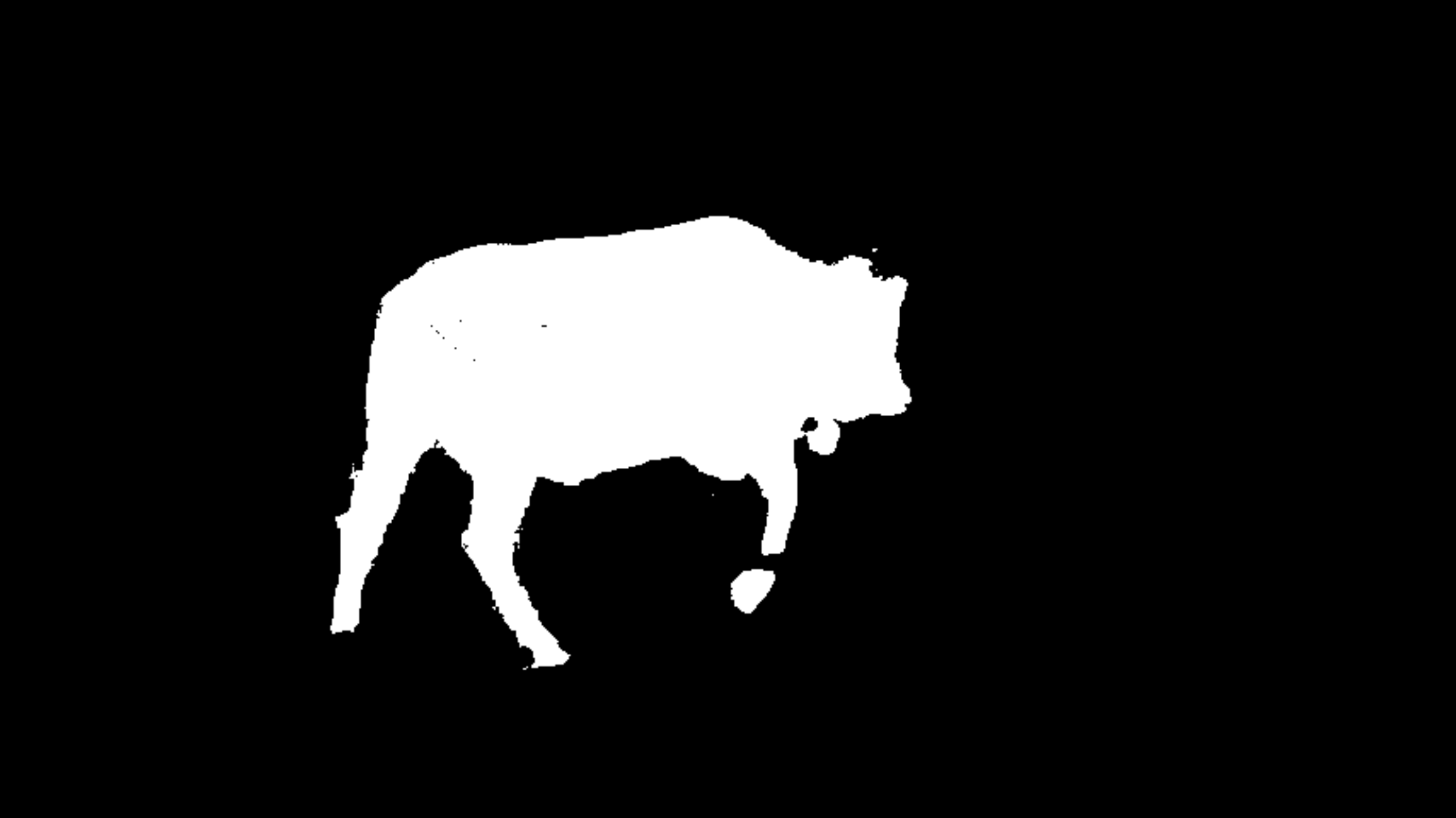}\\

            \includegraphics[width=0.12\textwidth]{davis_2017_val_00009.pdf}& 
            \includegraphics[width=0.12\textwidth]{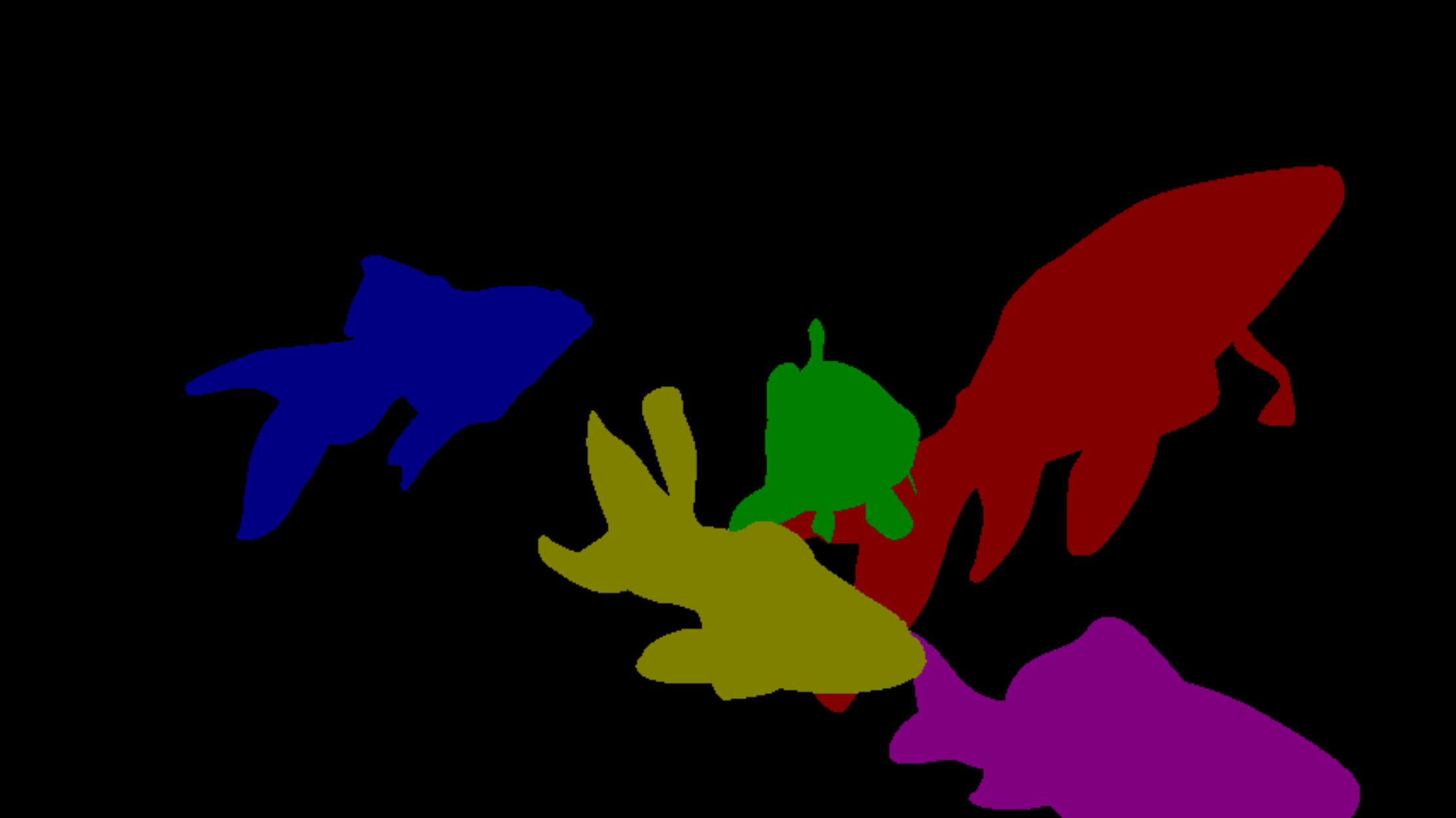}& 
            \includegraphics[width=0.12\textwidth]{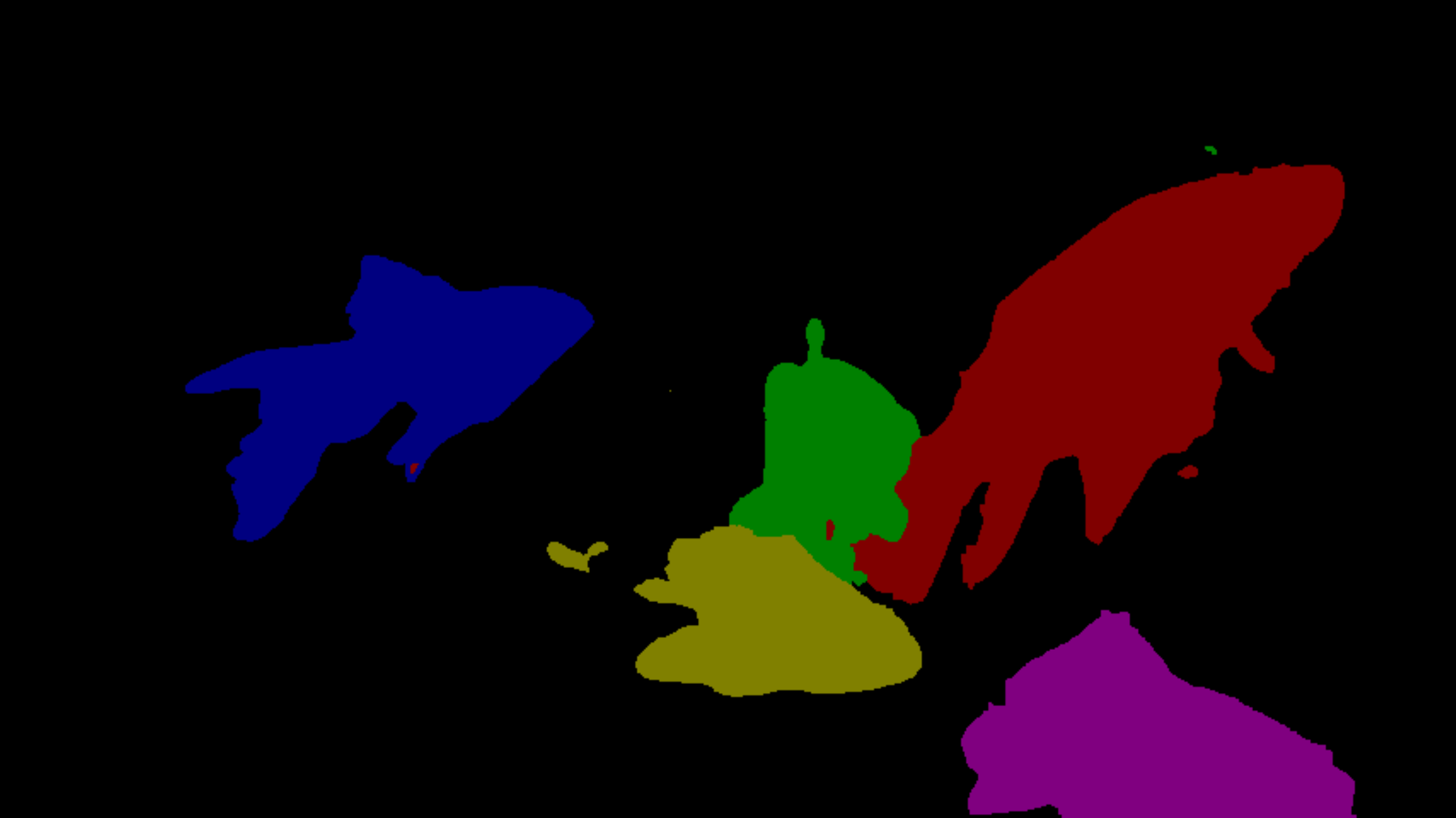}& 
            \includegraphics[width=0.12\textwidth]{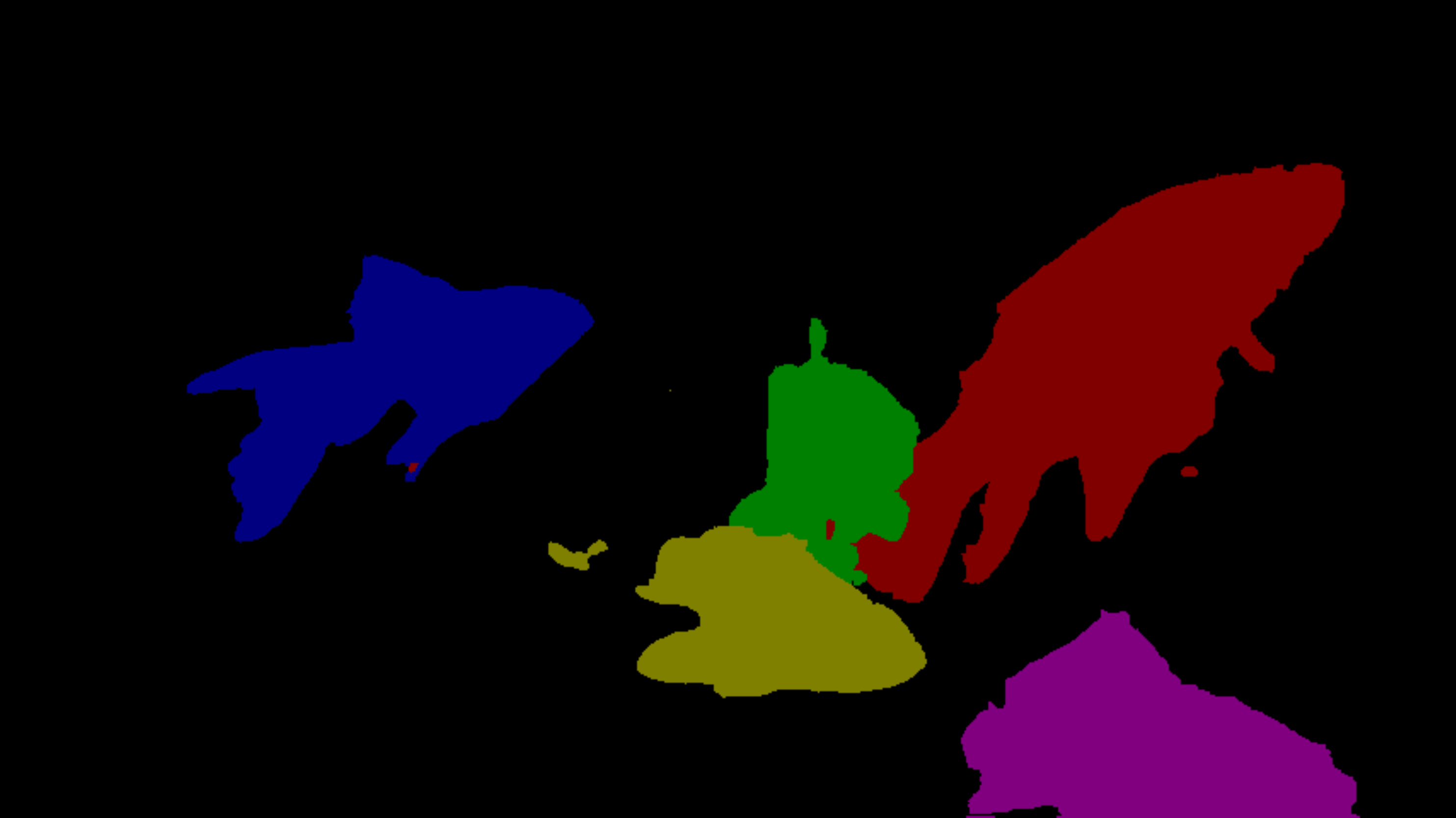}&
            \includegraphics[width=0.12\textwidth]{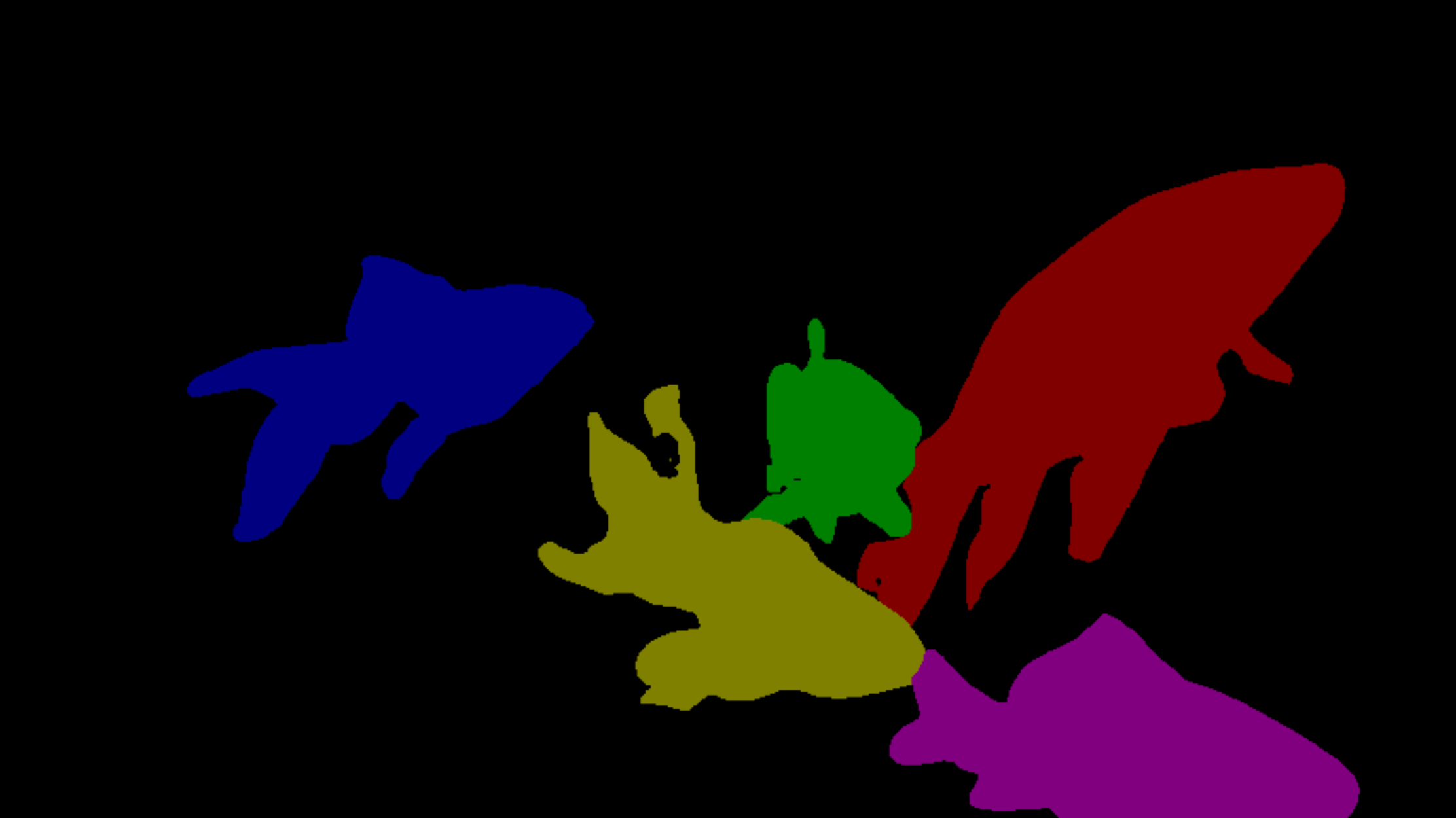}& 
            \includegraphics[width=0.12\textwidth]{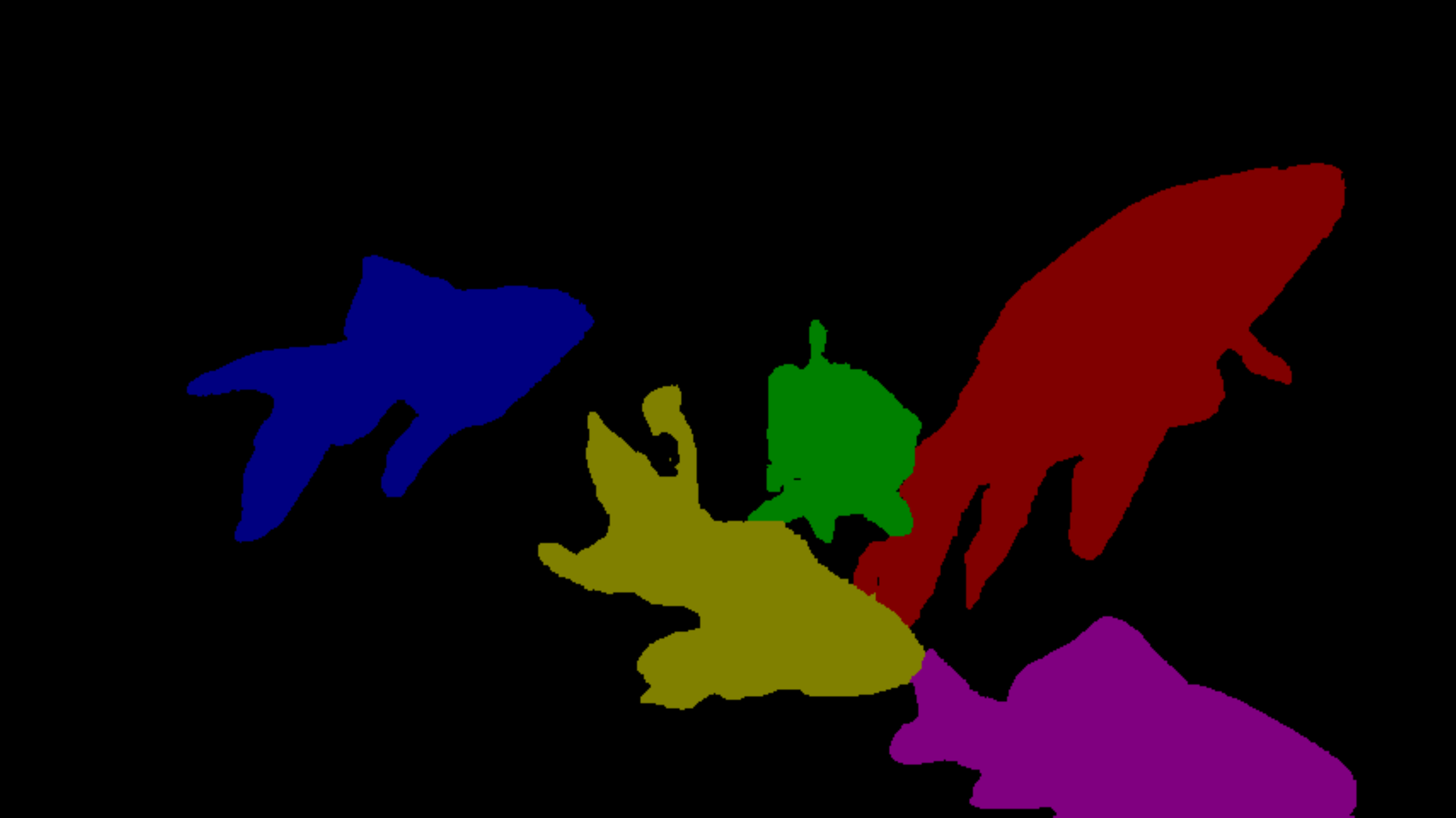}&
            \includegraphics[width=0.12\textwidth]{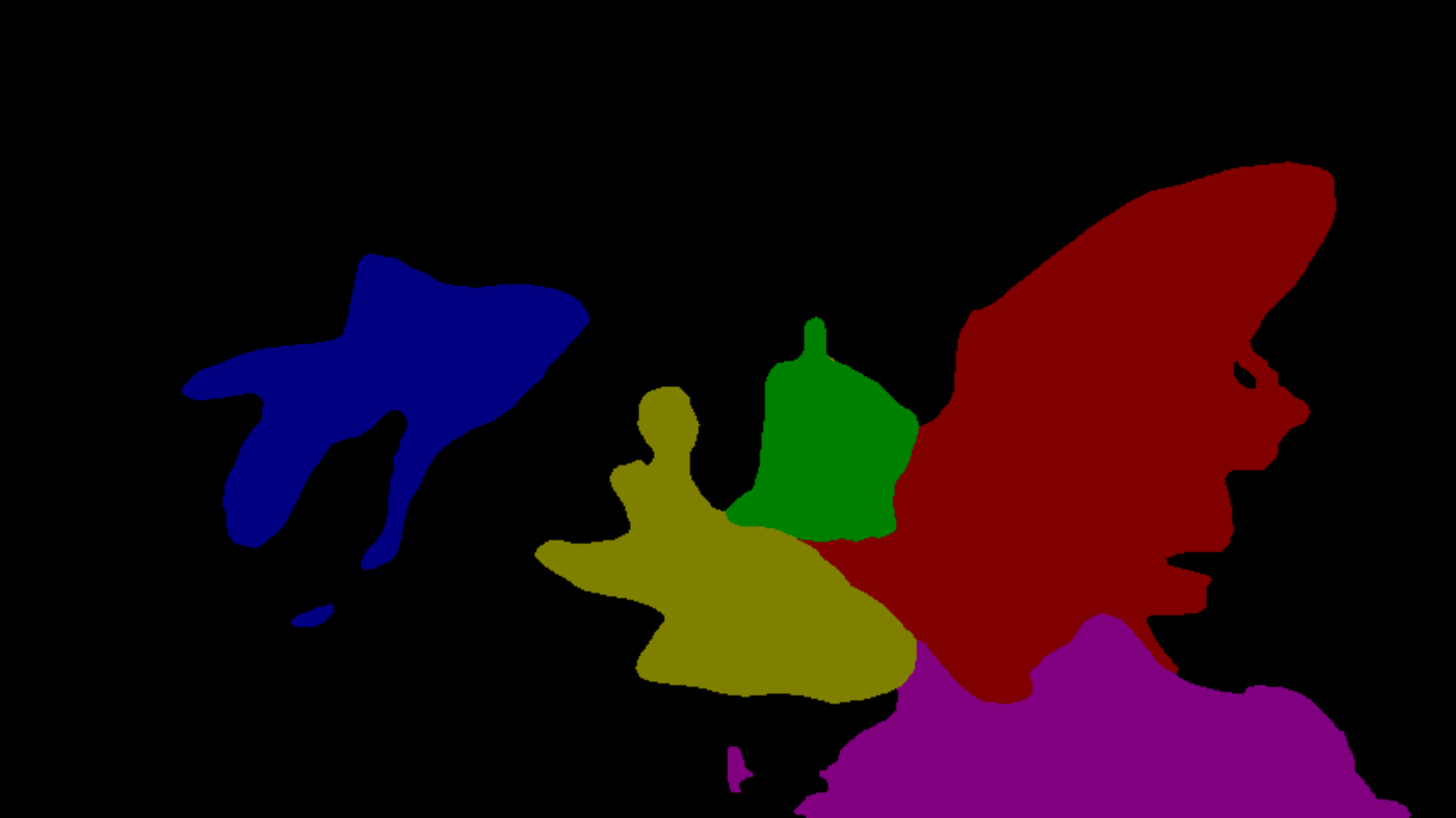}&
            \includegraphics[width=0.12\textwidth]{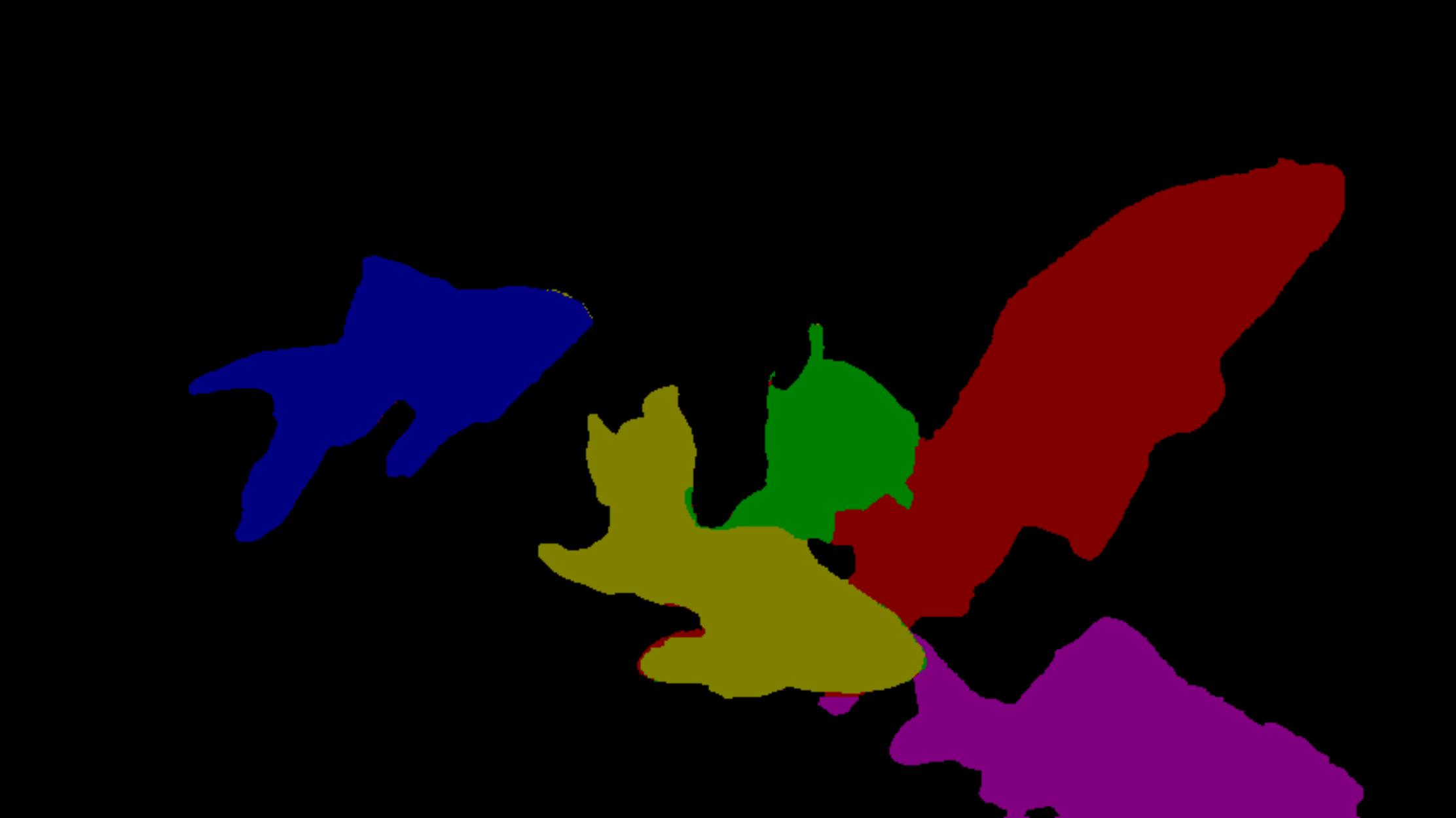}\\
            
            \includegraphics[width=0.12\textwidth]{loading_00030.pdf}& 
            \includegraphics[width=0.12\textwidth]{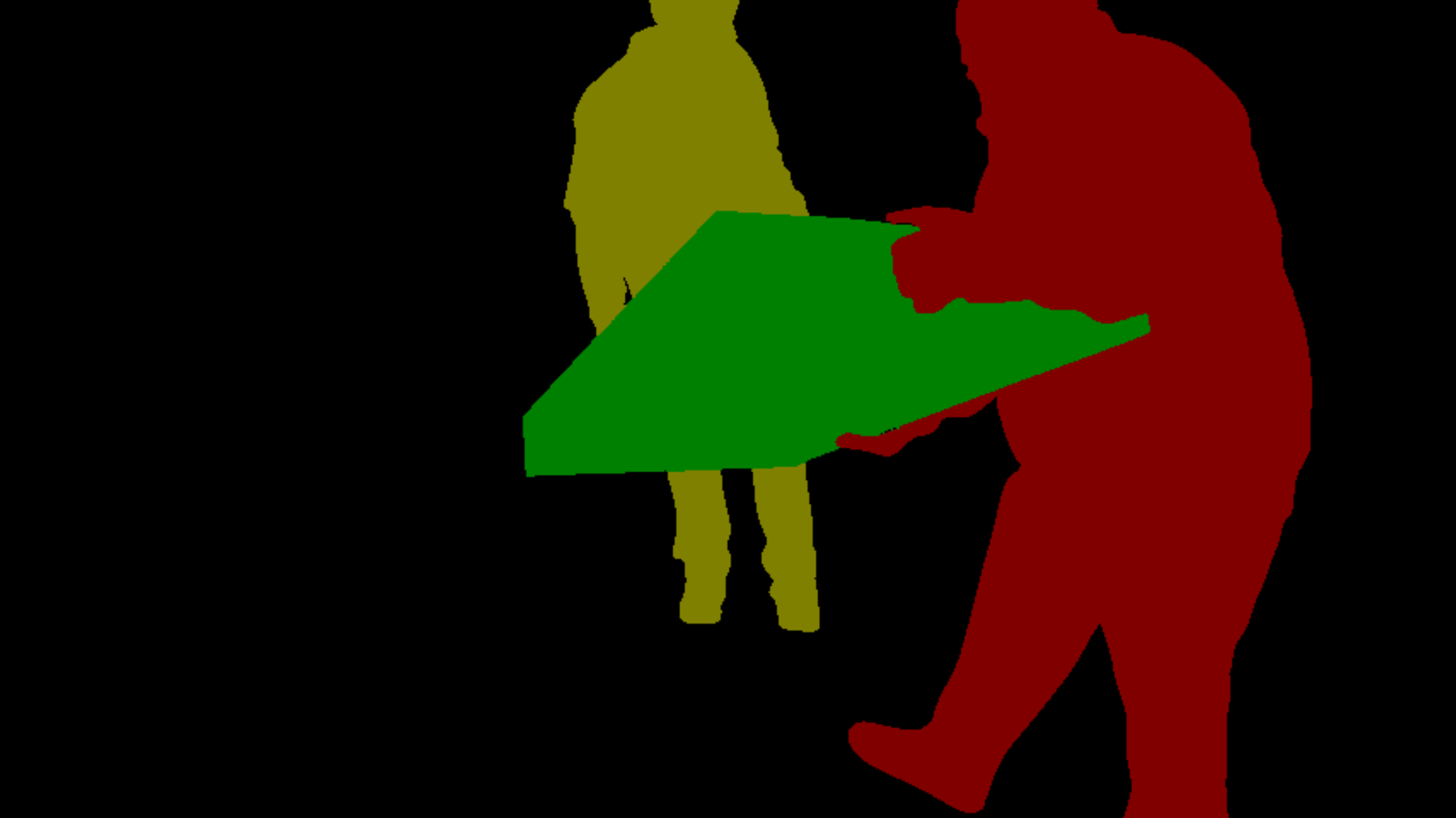}& 
            \includegraphics[width=0.12\textwidth]{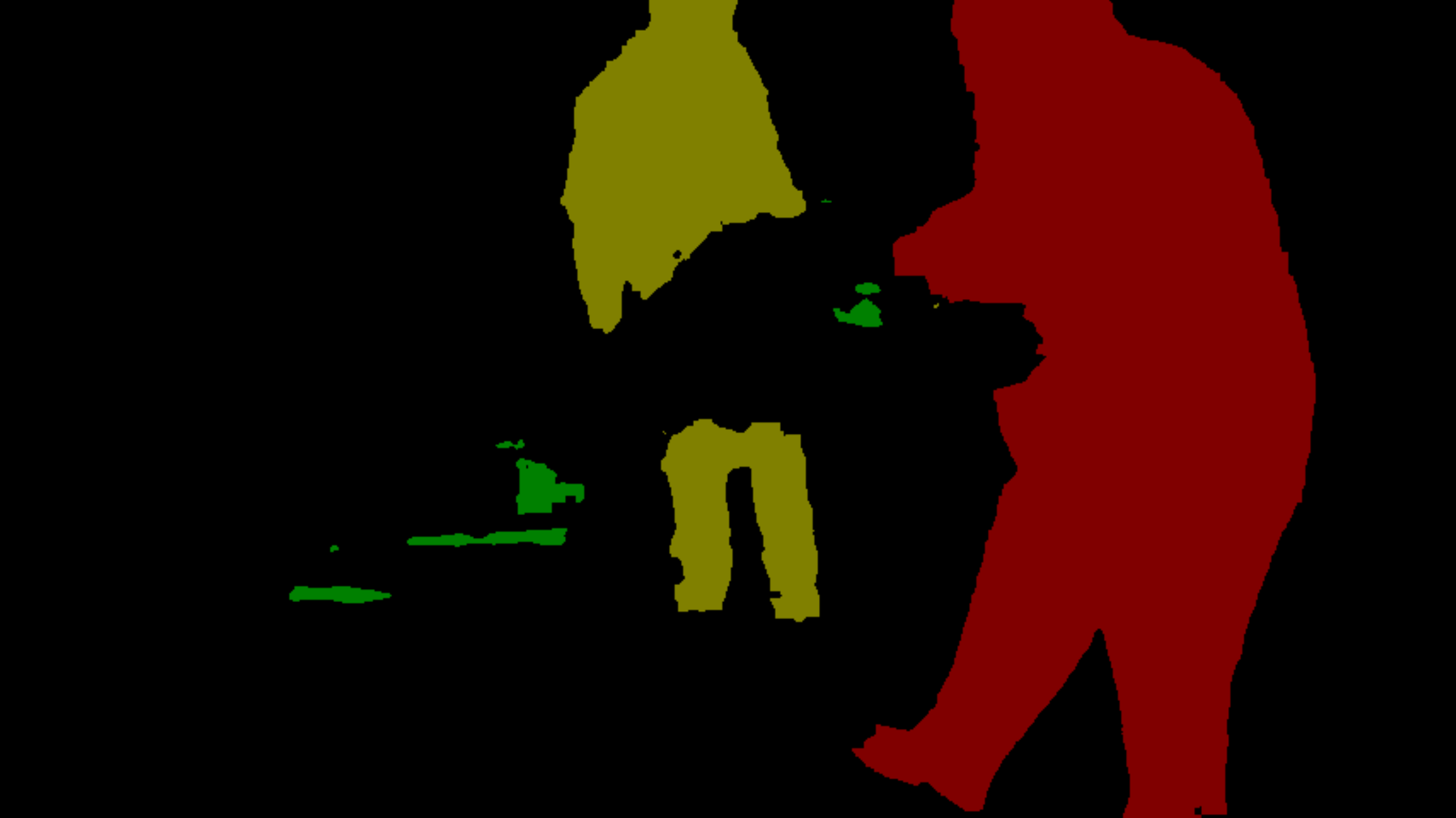}& 
            \includegraphics[width=0.12\textwidth]{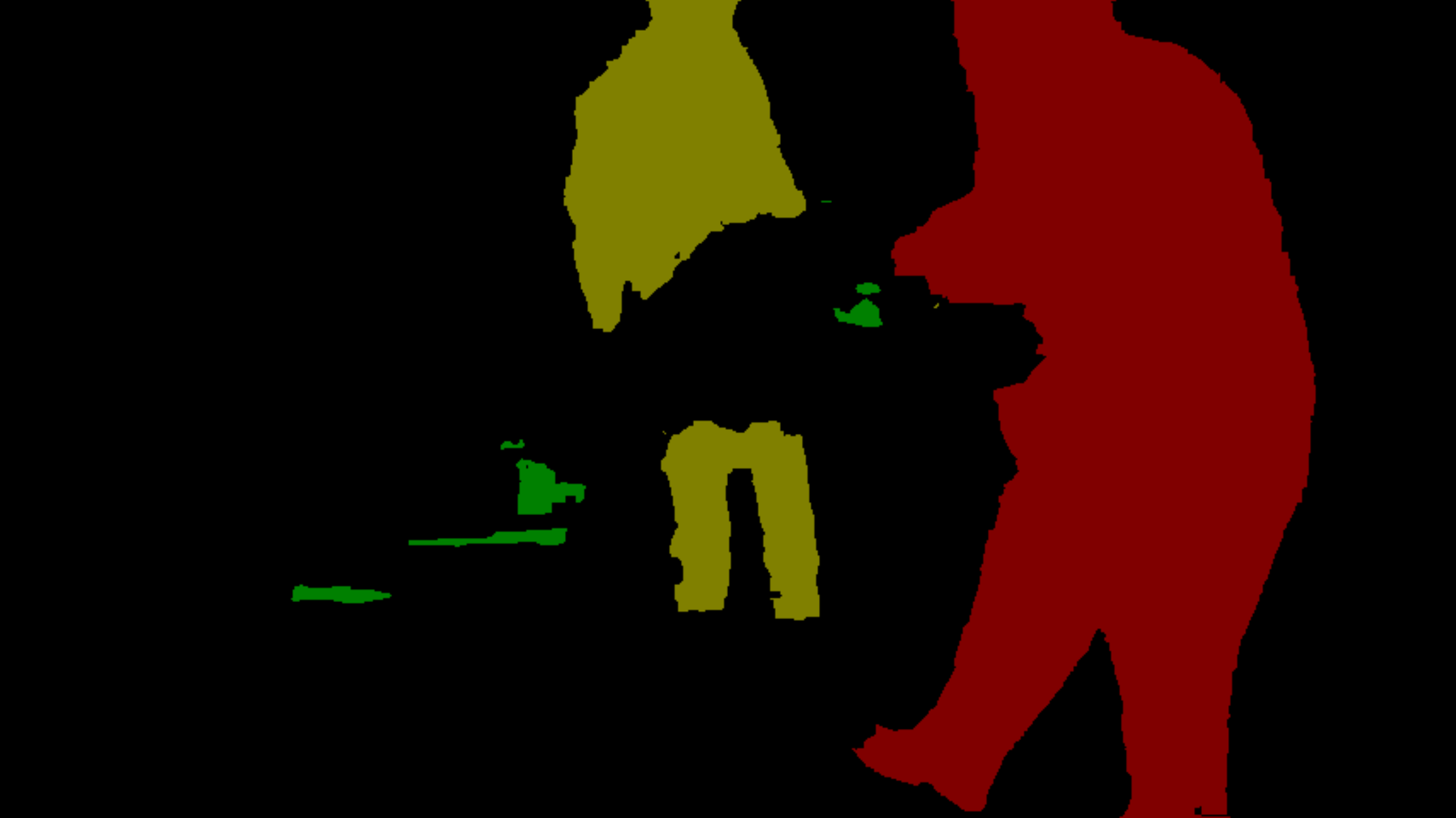}&
            \includegraphics[width=0.12\textwidth]{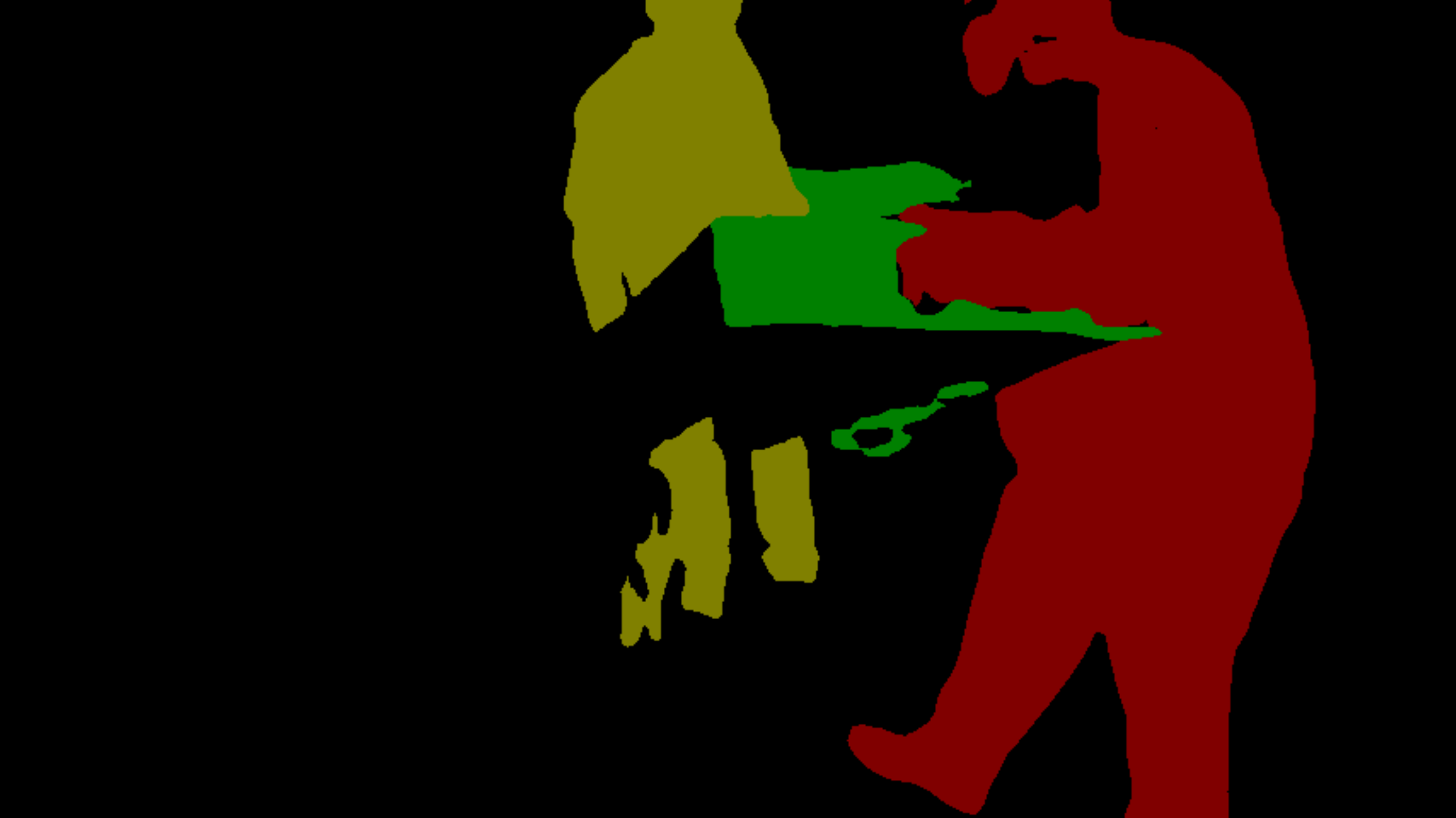}& 
            \includegraphics[width=0.12\textwidth]{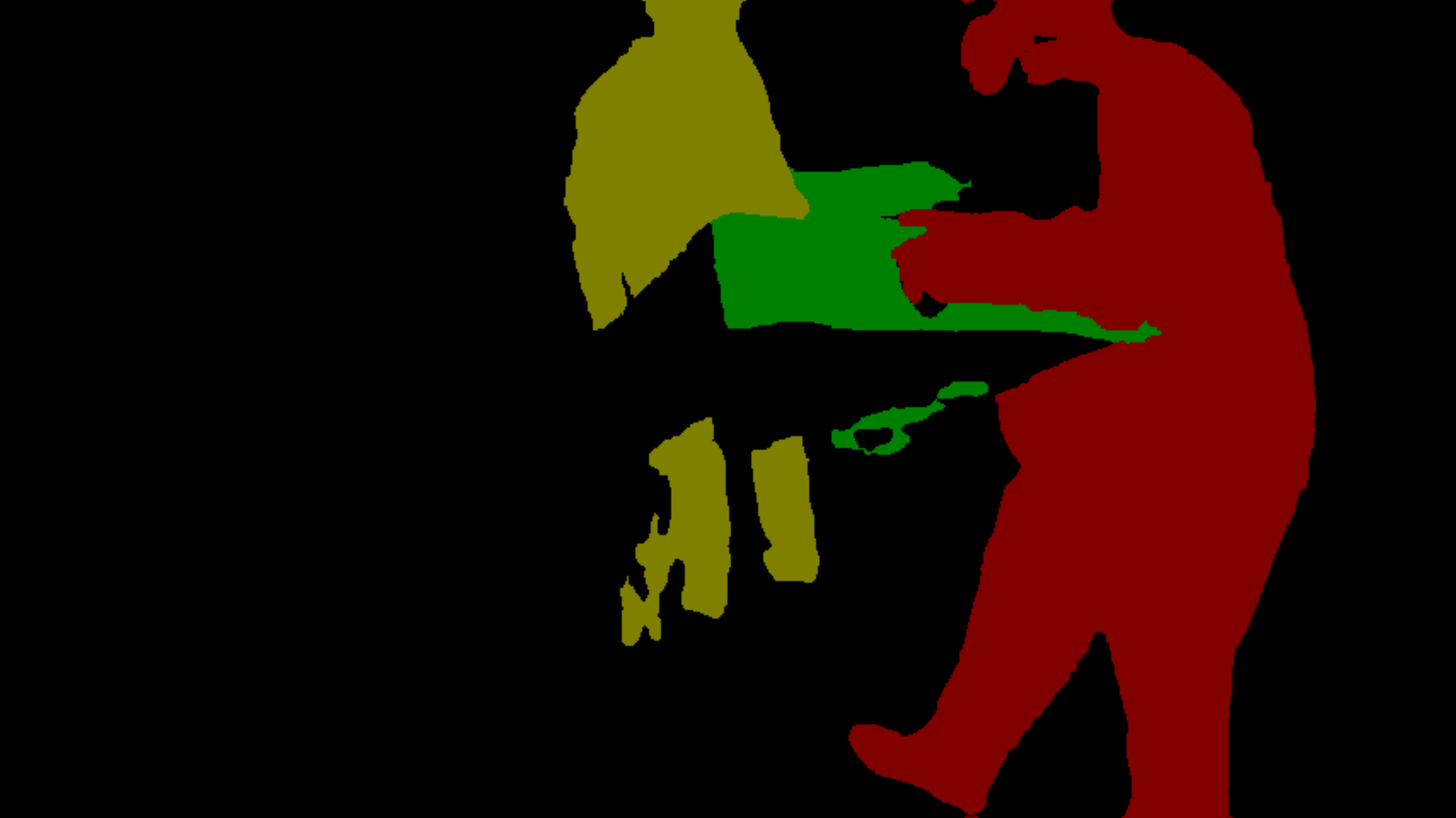}&
            \includegraphics[width=0.12\textwidth]{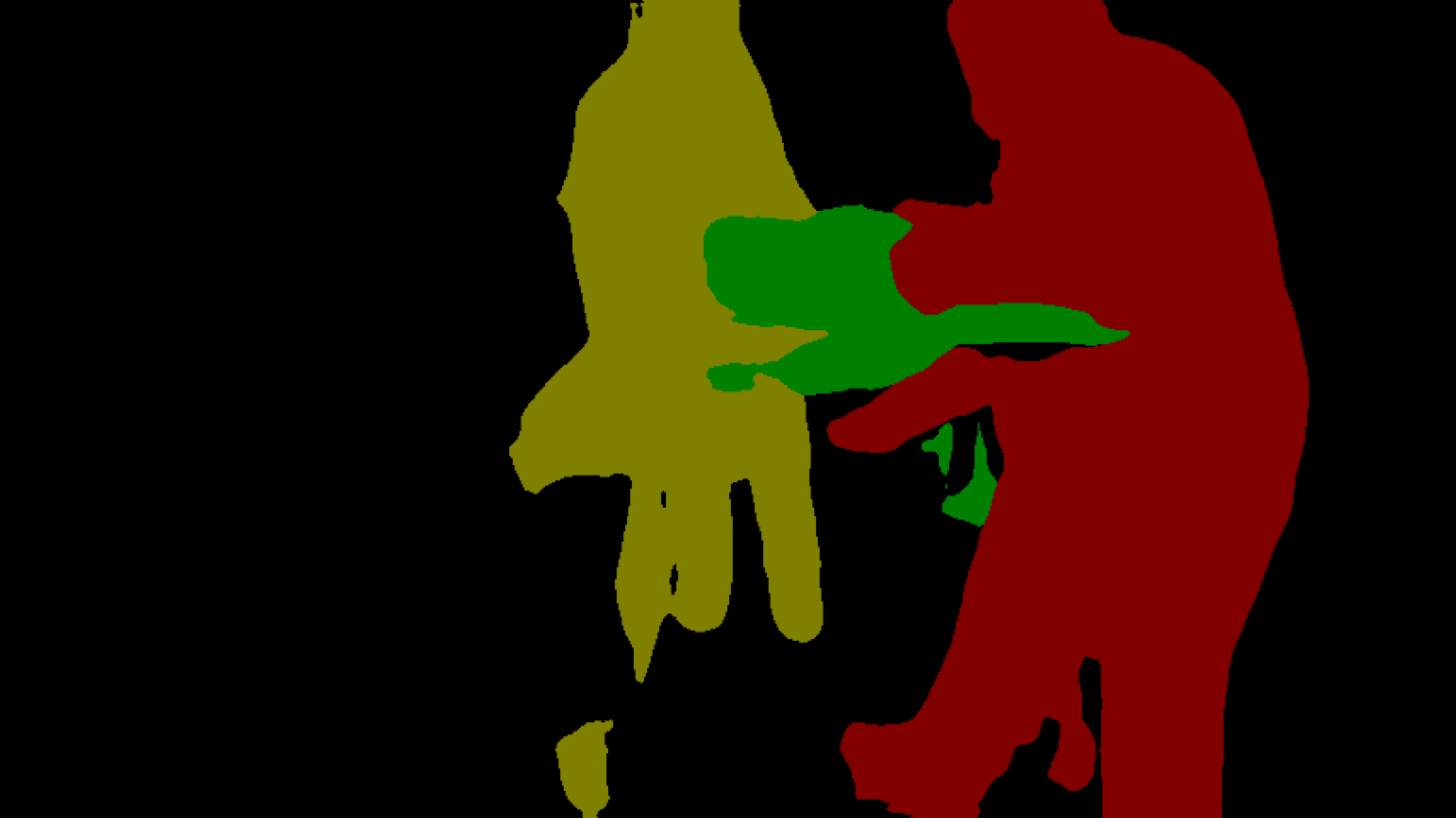}&
            \includegraphics[width=0.12\textwidth]{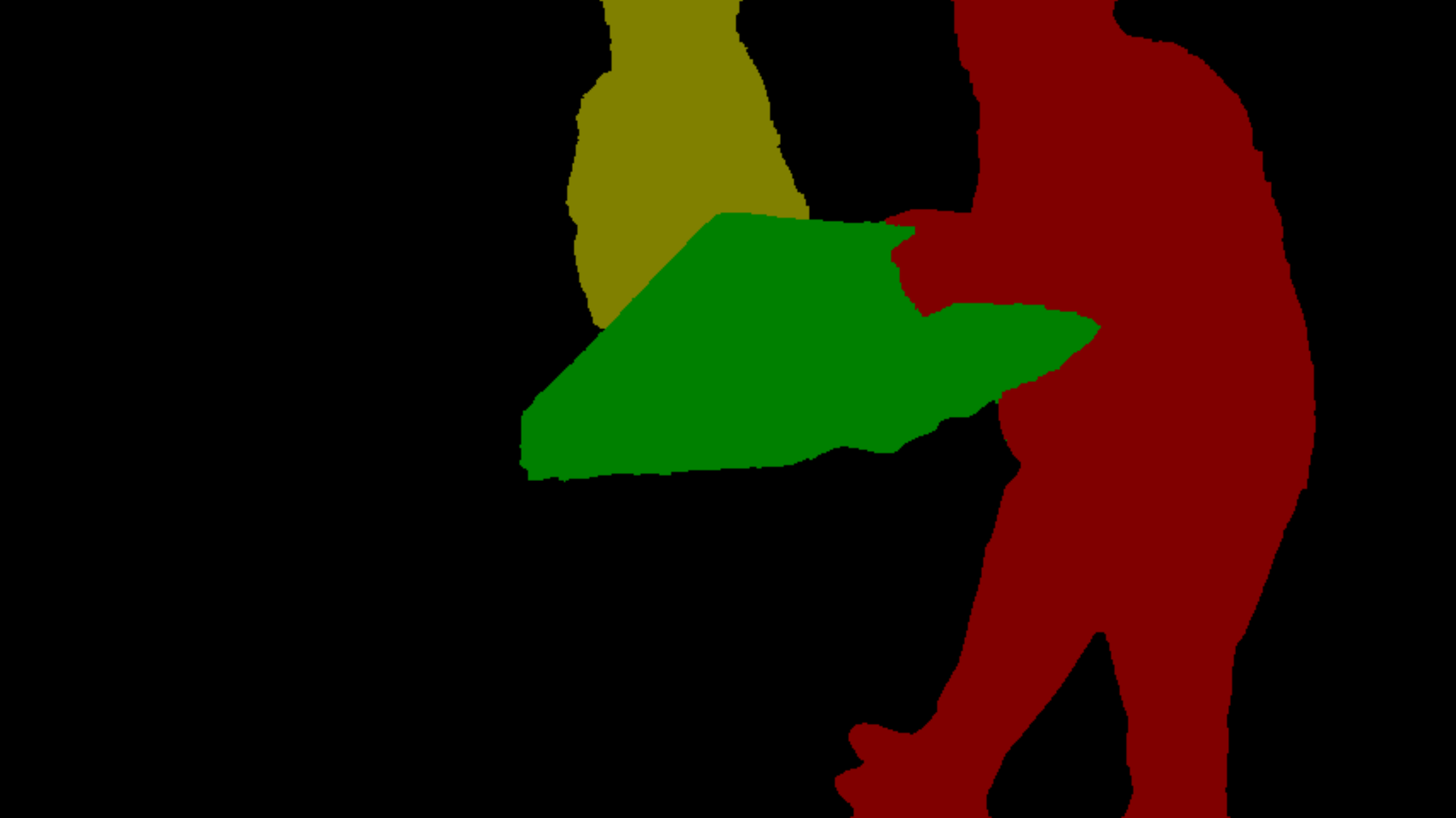} \\
            
            \includegraphics[width=0.12\textwidth]{pigs_00046.pdf}& 
            \includegraphics[width=0.12\textwidth]{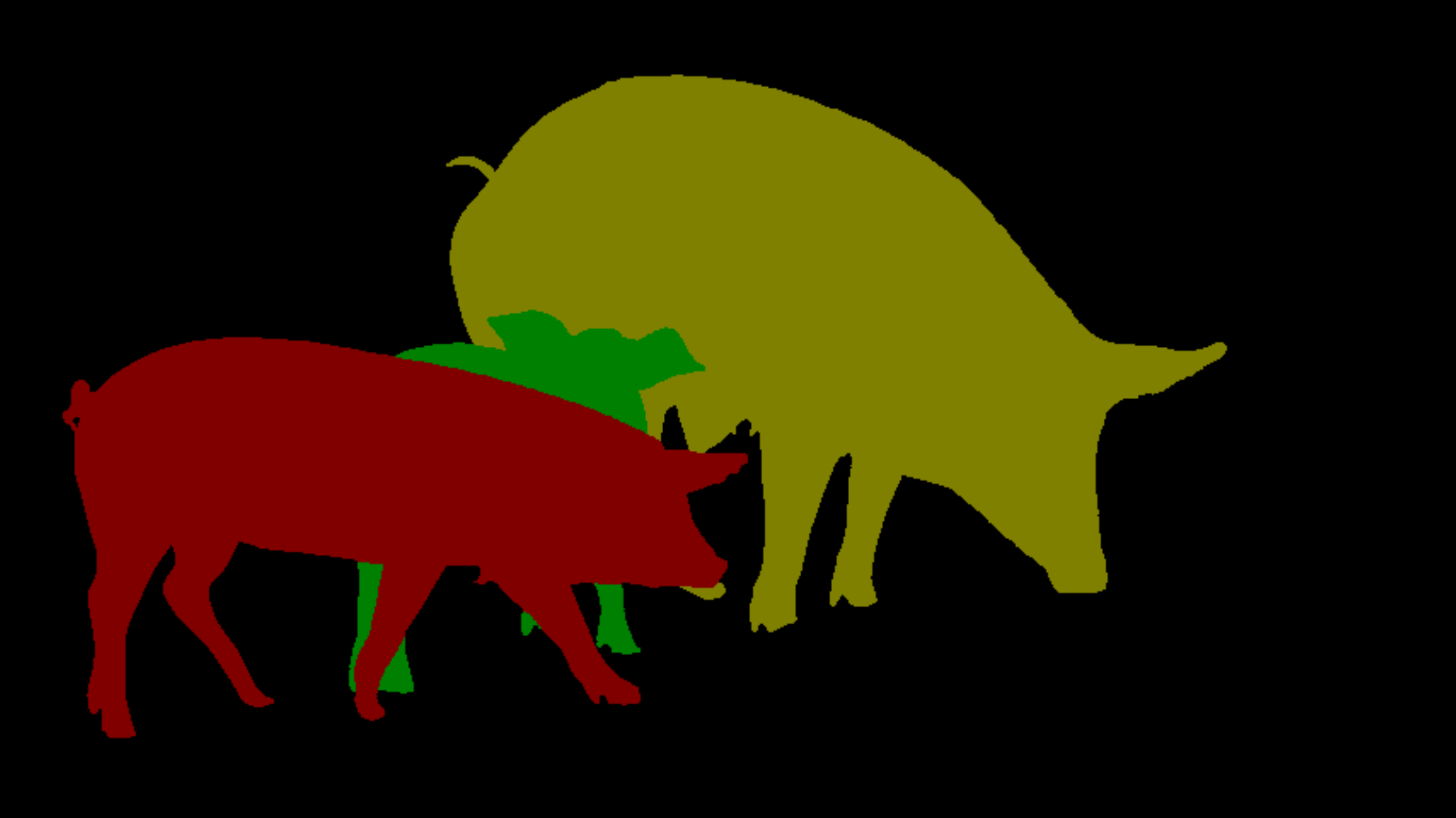}& 
            \includegraphics[width=0.12\textwidth]{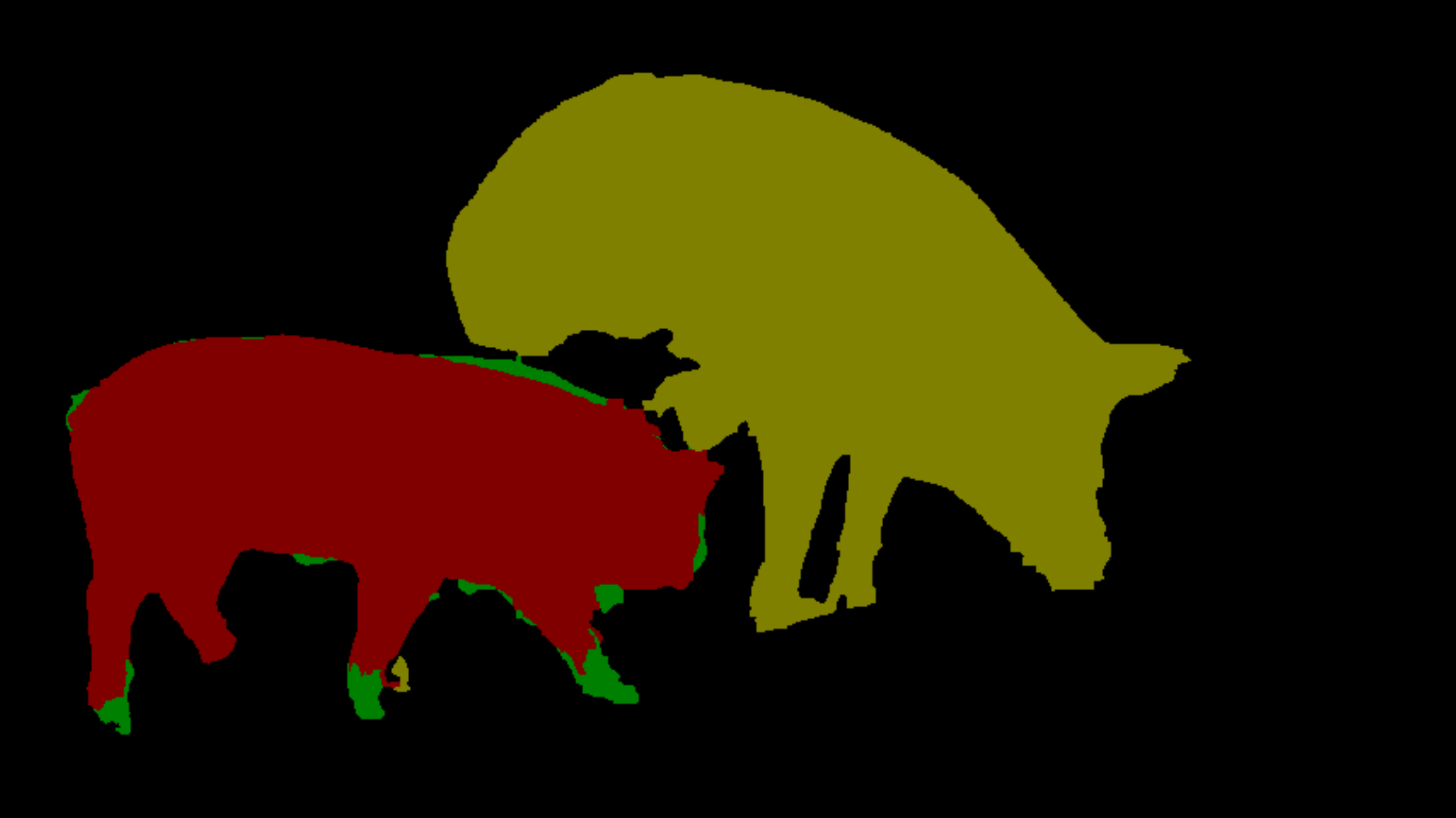}&
            \includegraphics[width=0.12\textwidth]{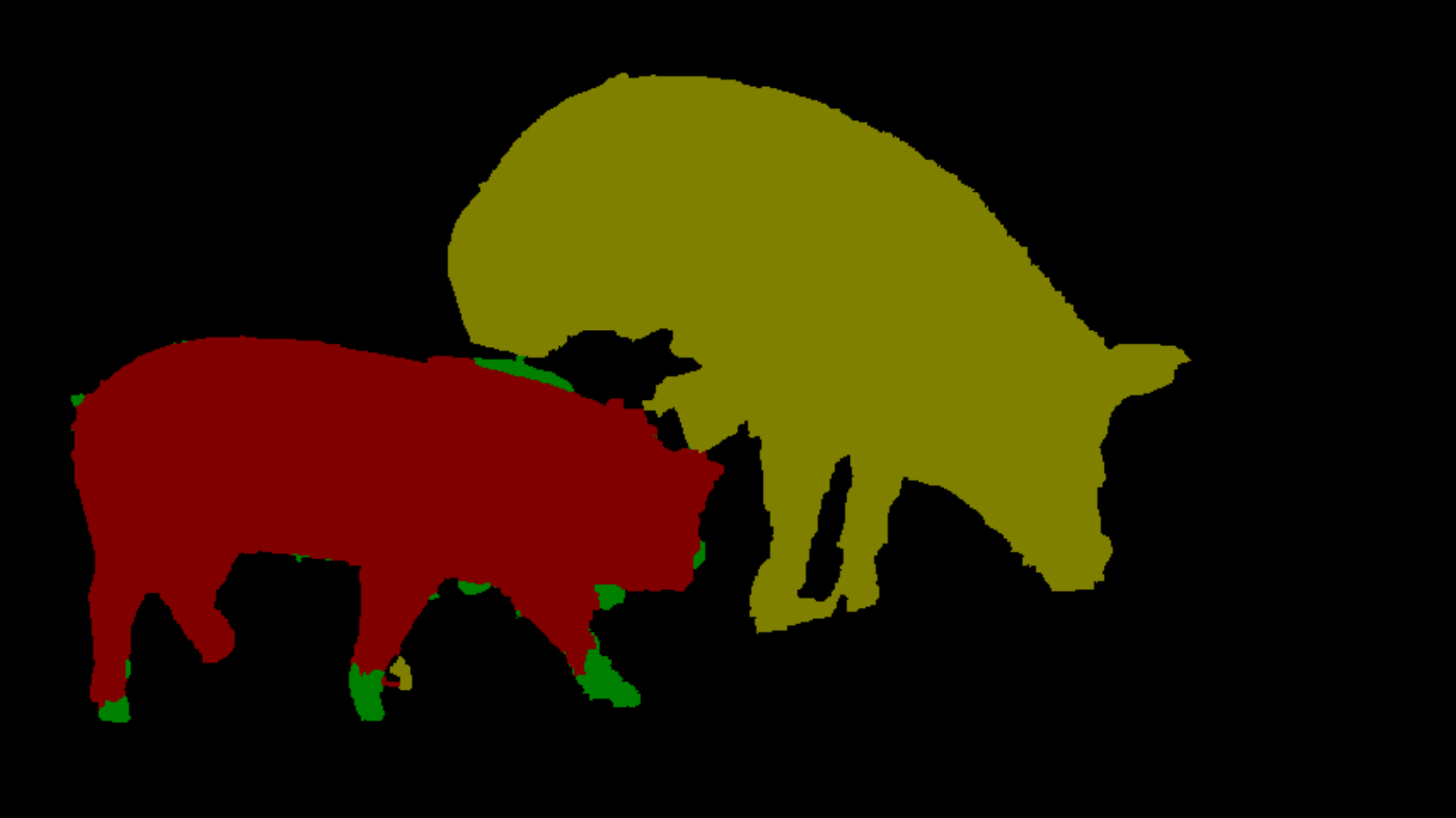}&
            \includegraphics[width=0.12\textwidth]{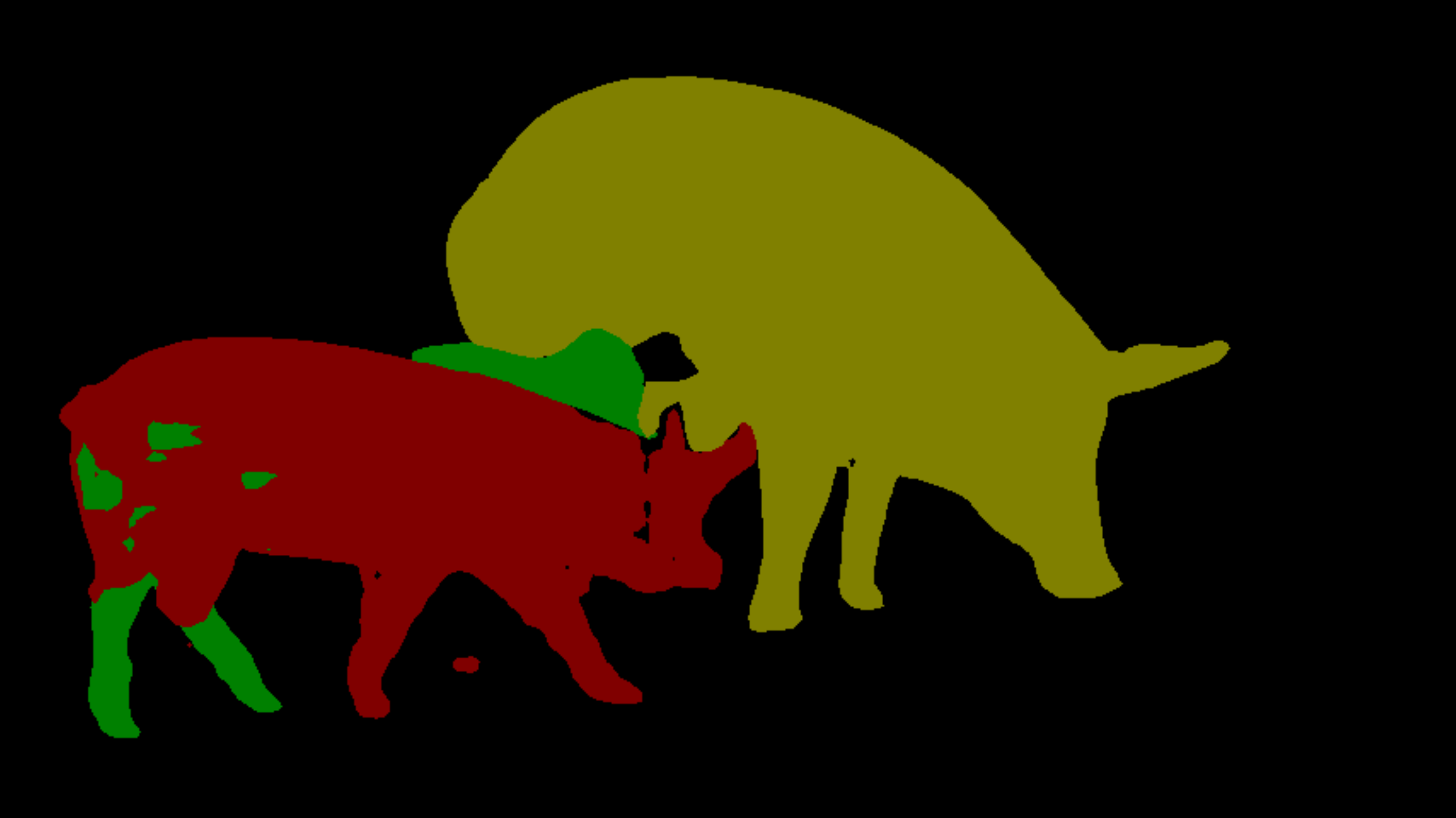}& 
            \includegraphics[width=0.12\textwidth]{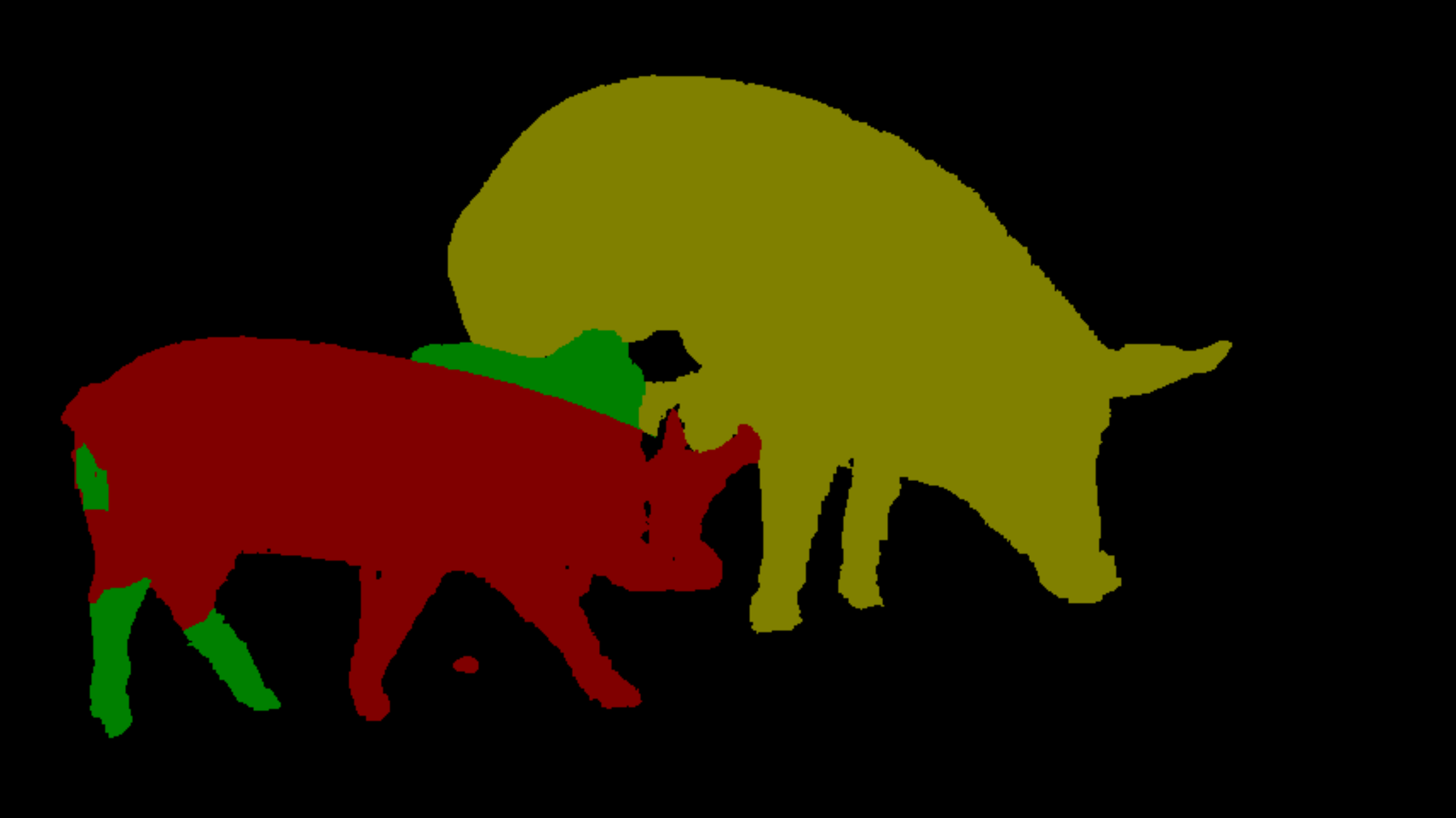}&
            \includegraphics[width=0.12\textwidth]{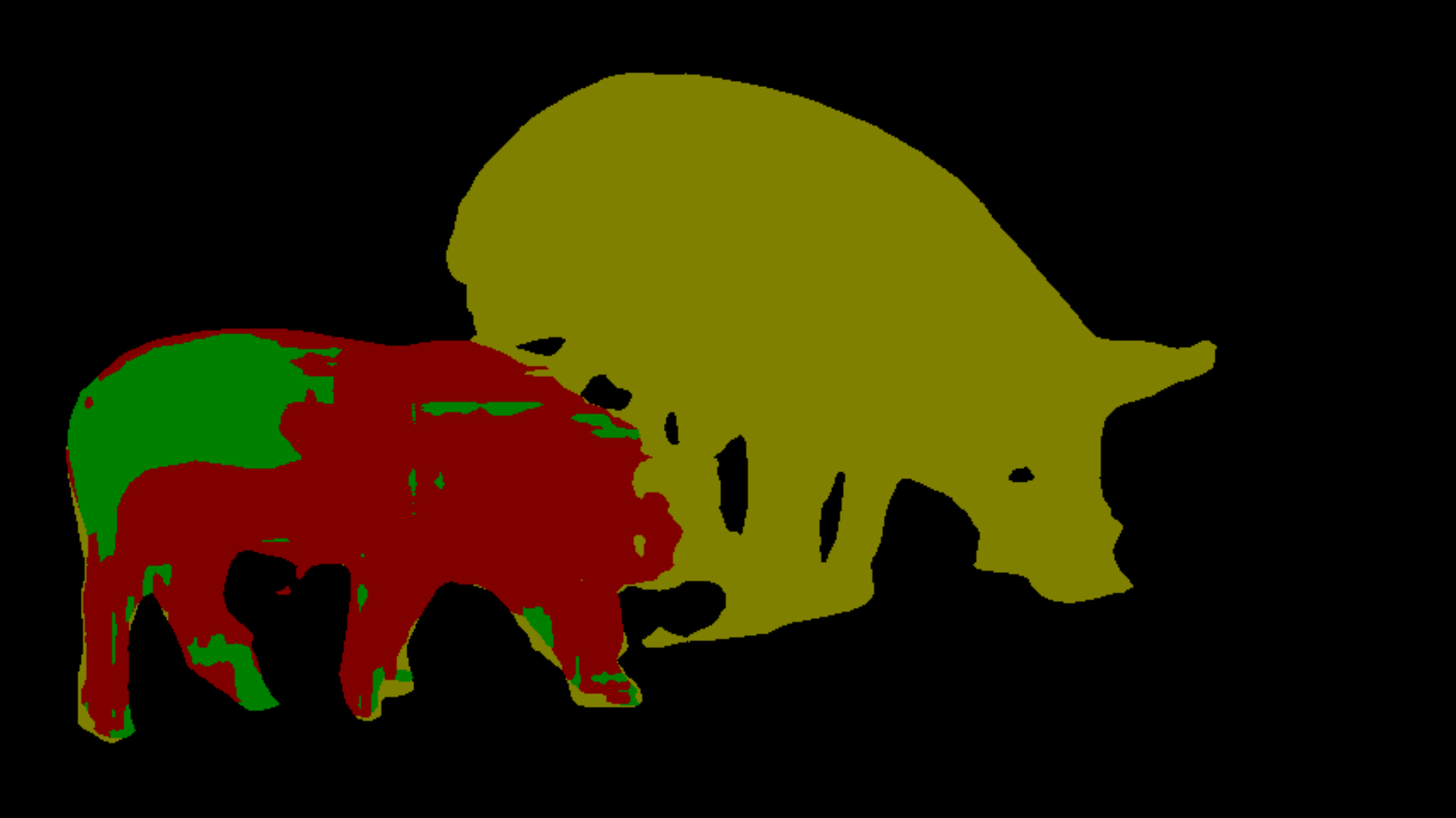}&
            \includegraphics[width=0.12\textwidth]{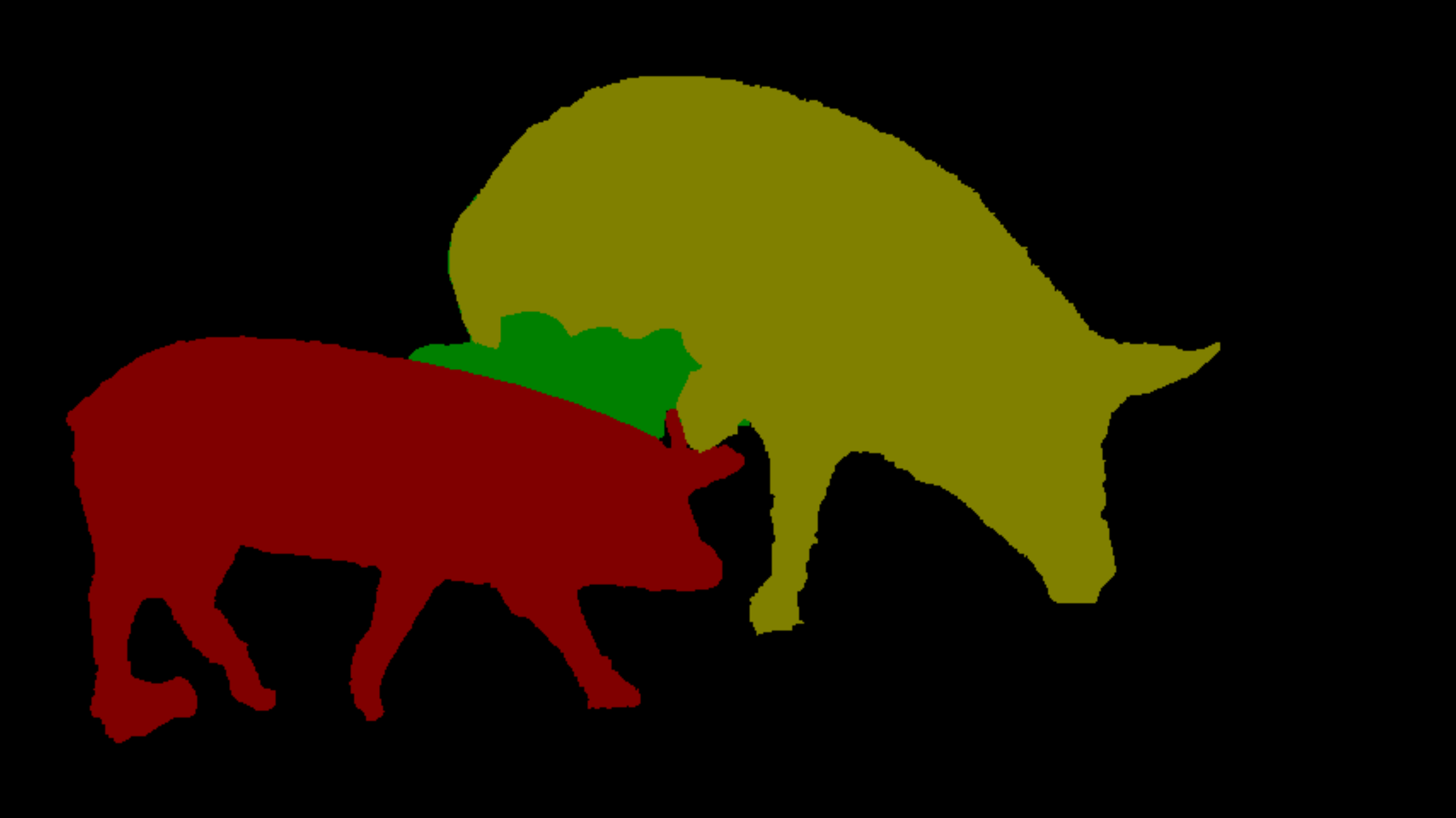} \\
            
            image & GT & \cite{c21} & \cite{c21} + $ours$ & \cite{c20} & \cite{c20} + $ours$ & \cite{c27} & $ours$

\end{tabular} 
		
	\caption{Exemplary results for segmentation tracking on the DAVIS$_{16}$ (binary) and DAVIS$_{17}$ (multi-label) benchmark. We compare different state-of-the-art methods (OSVOS-S~\cite{c21}, CINM~\cite{c20} and OSMN~\cite{c27}) and {\bf ours}. 
	}
\label{results_2017}
\end{figure*}

\subsection{Results on DAVIS}

\begin{table}[t!]
		\centering
	\caption{Results on train and validation sets of DAVIS$_{16}$. We report Mean (M), Recall (R) and Decay (D) of the evaluation metrics ($\it{F}$ and $\it{J}$).}
		\label{davis_bench}
	\begin{tabular}{@{}c@{}c@{\hspace{0.15cm}}|@{\hspace{0.15cm}}c@{\hspace{0.2cm}}c@{\hspace{0.2cm}}c@{\hspace{0.15cm}}|@{\hspace{0.15cm}}c@{\hspace{0.2cm}}c@{\hspace{0.2cm}}c@{}}
	\toprule
	 \multicolumn{8}{@{}c@{}}{\textbf{train}} \\
	\midrule
	 &Measure &  \multicolumn{3}{@{}c@{}}{$\it{F(\%)}$} &  \multicolumn{3}{@{}c@{}}{$\it{J(\%)}$} \\
 	& & M & R & D & M & R & D\\
 	\midrule
    \parbox[t]{2mm}{\multirow{3}{*}{\rotatebox[origin=c]{90}{CNN}}}
 	& MSK~\cite{c2} & $76.1$ & $88.9$ & $9.8$ & $\mathbf{80.7}$ & $93.9$  & $8.8$ \\
 	& VPN~\cite{jampani2016video}   & $\mathbf{77.0}$ & $\mathbf{94.3}$ & $13.1$ & $78.3$ & $\mathbf{95.4}$  & $7.2$ \\
 	& CTN~\cite{CTN} & $72.8$ & $88.3$ & $14.7$ & $76.9$ & $90.0$ & $13.5$ \\
 	\midrule
 	\parbox[t]{2mm}{\multirow{5}{*}{\rotatebox[origin=c]{90}{non-CNN}}}
 	& OFL~\cite{c18} & $70.9$ & $83.1$ & $21.9$ & $73.2$ & $83.0$  & $20.2$\\
 	& BVS~\cite{c29} & $70.1$ & $83.7$ & $25.1$ & $70.9$ & $82.7$  & $24.1$ \\
 	& FCP~\cite{c30} & $58.3$ & $67.6$ & $\mathbf{7.2}$ & $66.2$ & $82.0$  & $\mathbf{6.5}$ \\
 	& JMP~\cite{c31} & $62.3$ & $73.2$ & $36.5$ & $63.2$ & $73.7$  & $35.8$ \\
 	& \emph{ours} & $\mathbf{73.5}$ & $\mathbf{88.6}$ & $13.8$ & $\mathbf{77.7}$ & $\mathbf{90.2}$  & $12.5$ \\
 	\bottomrule
 	\toprule
 	\multicolumn{8}{@{}c@{}}{\textbf{validation}} \\
 	\midrule
 	& Measure &  \multicolumn{3}{@{}c@{}}{$\it{F(\%)}$} &  \multicolumn{3}{@{}c@{}}{$\it{J(\%)}$} \\
 	& & M & R & D & M & R & D\\
 	\midrule
 	\parbox[t]{2mm}{\multirow{9}{*}{\rotatebox[origin=c]{90}{CNN}}}
 	& OSVOS-S~\cite{c21} & $87.5$ & $\mathbf{95.9}$ & $8.2$ & $85.6$ & $96.8$  & $5.5$\\
 	& \cite{c21} + \emph{ours} & $87.6$ & $\mathbf{95.9}$ & $8.1$ & $\mathbf{86.0}$ & $\mathbf{96.9}$ & $5.6$ \\
 	& CINM~\cite{c20}    & $85.0$ & $92.1$ & $14.7$ & $83.4$ & $94.9$ & $12.3$ \\
 	& \cite{c20} + \emph{ours}    & $\mathbf{87.7}$ & $93.0$ & $14.3$ & $84.2$ & $95.6$ & $12.1$ \\ 
 	& MSK~\cite{c2} & $75.4$ & $87.1$ & $9.0$ & $79.7$ & $93.1$  & $8.9$ \\
 	& VPN~\cite{jampani2016video}  & $65.5$  &  $69.0$  &  $14.4$ & $70.2$ &  $82.3$ &  $12.4$ \\
 	& SIAMMASK~\cite{siammask}  & $67.8$  &  $79.8$ & $2.1$  & $71.7$ & $86.8$ & $3.0$  \\
 	& CTN~\cite{CTN}  & $69.3$ & $79.6$ & $12.9$ & $73.5$ & $87.4$ & $15.6$   \\
 	& PLM~\cite{c28} & $62.5$ & $73.2$ & $14.7$ & $70.2$ & $86.3$  & $11.2$ \\
 	\midrule
 	\parbox[t]{2mm}{\multirow{5}{*}{\rotatebox[origin=c]{90}{non-CNN}}}
 	& OFL~\cite{c18} & $63.4$ & $70.4$ & $27.2$ & $68.0$ & $75.6$  & $26.4$ \\
 	& BVS~\cite{c29} & $58.8$ & $67.9$ & $21.3$ & $60.0$ & $66.9$  & $28.9$ \\
 	& FCP~\cite{c30} & $49.2$ & $49.5$ & $\mathbf{-1.1}$ & $58.4$ & $71.5$  & $\mathbf{-2.0}$ \\
 	& JMP~\cite{c31} & $53.1$ & $54.2$ & $38.4$ & $57.0$ & $62.6$  & $39.4$ \\
 	& \emph{ours} & $\mathbf{68.4}$ & $\mathbf{78.4}$ & $17.8$ & $\mathbf{71.6}$ & $\mathbf{81.0}$  & $16.8$ \\
 	\bottomrule
	\end{tabular}
	\label{davis_bench2}
\end{table}

\begin{table}[t]
 		\centering
 		\caption{Results on the DAVIS$_{17}$ validation and test set. 
 		} 
 		\label{dataterm_2017}
 		\begin{tabular}{l|@{\hspace{0.3cm}} c @{\hspace{0.3cm}} c @{\hspace{0.3cm}}   c @{\hspace{0.3cm}} c}
 		\toprule
 			&\multicolumn{2}{c}{$\it{F(\%)}$}	& \multicolumn{2}{c}{$\it{J(\%)}$}        \\ \midrule
 			Method 		& val & test  & val & test  \\ 
 			\midrule 
 			\emph{ours} 	& $56.5$  & $44.0$ &  $54.5$  &  $41.5$\\ 
 			\midrule
 			OSMN~\cite{c27} &  $57.1$ & $44.9$  & $ 52.5$ & $37.7$\\
 			FAVOS\cite{c32} &  $61.8$ & $44.2$  & $ 54.6$ & $42.9$\\
			OSVOS-S~\cite{c21}	&  $71.3$ & $62.1$  & $64.7$ & $52.9$ \\
			\cite{c21} + \emph{ours} &   $71.4$	& $62.2$  & $65.2$ & $53.7$ \\
			CINM ~\cite{c20} &  $74.0$ & $70.5$ & $67.2$ & $64.5$ \\
			\cite{c20} + \emph{ours} &  $\mathbf{74.1}$ & $\mathbf{70.6}$ & $\mathbf{67.4}$ & $\mathbf{64.7}$ \\
			\midrule
		\end{tabular}
\end{table}
	
\begin{figure}
	\begin{center}
	\small
	\begin{tabular}{@{}c@{}c@{}c@{}c@{}}
 
 \includegraphics[width=0.15\textwidth]{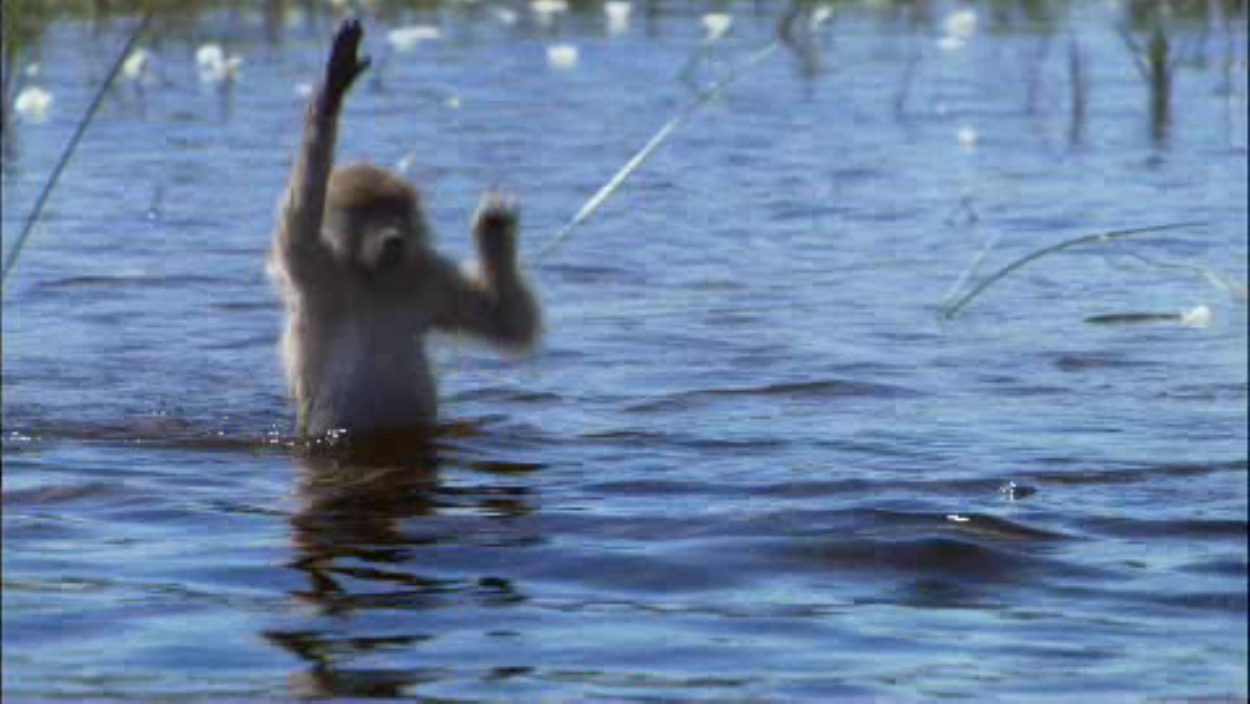}\,&
 \includegraphics[width=0.15\textwidth]{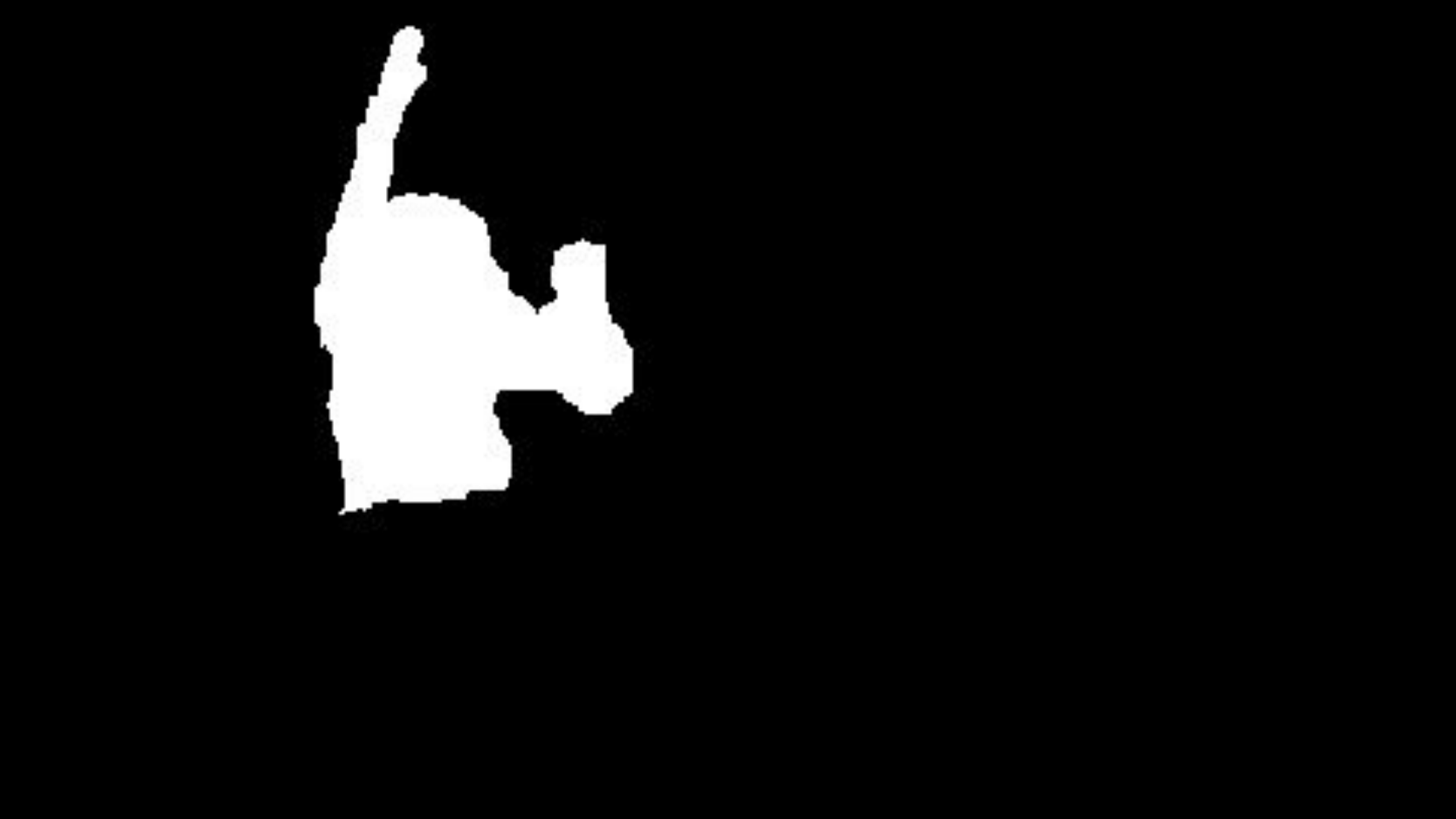}\,&
 \includegraphics[width=0.15\textwidth]{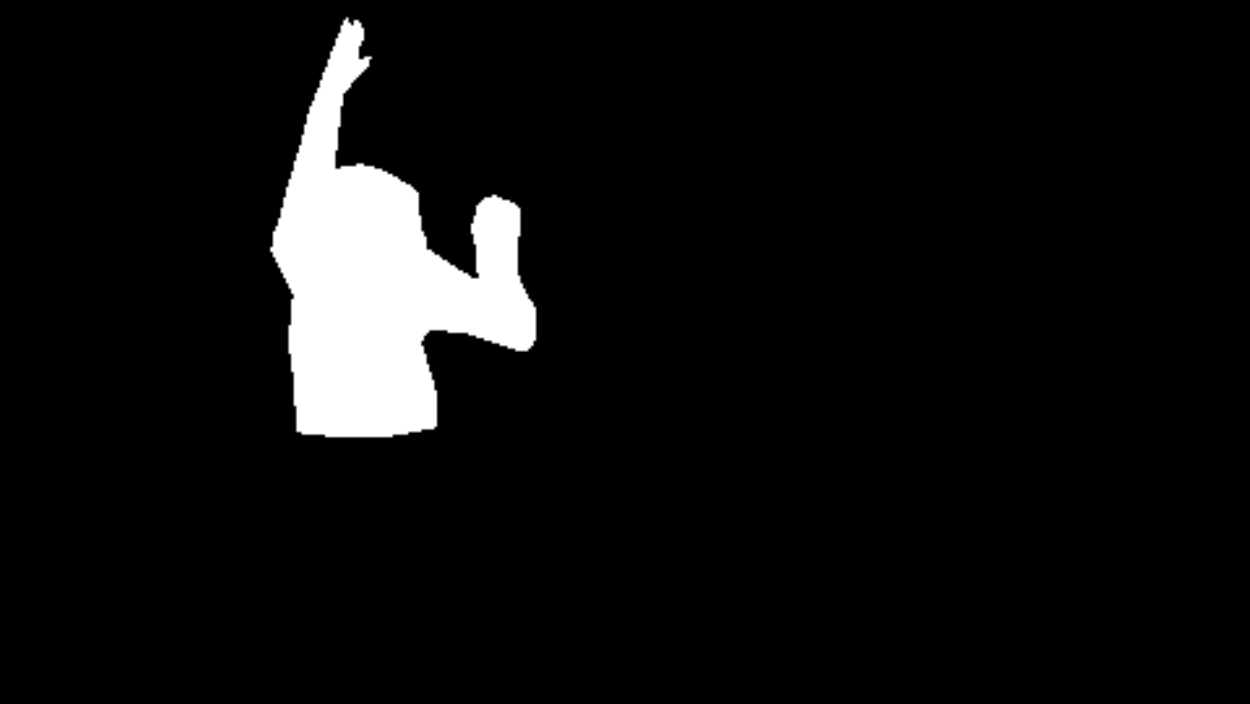}\,&\\
 
 \includegraphics[width=0.15\textwidth]{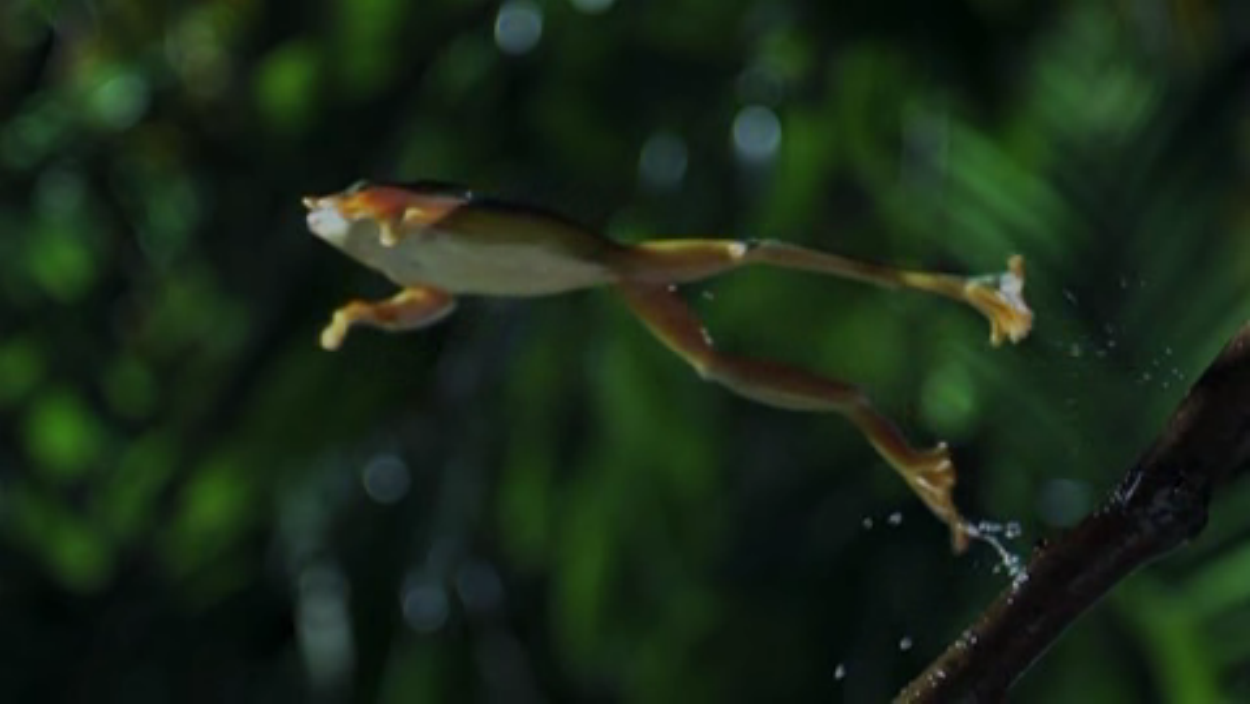}\,&
 \includegraphics[width=0.15\textwidth]{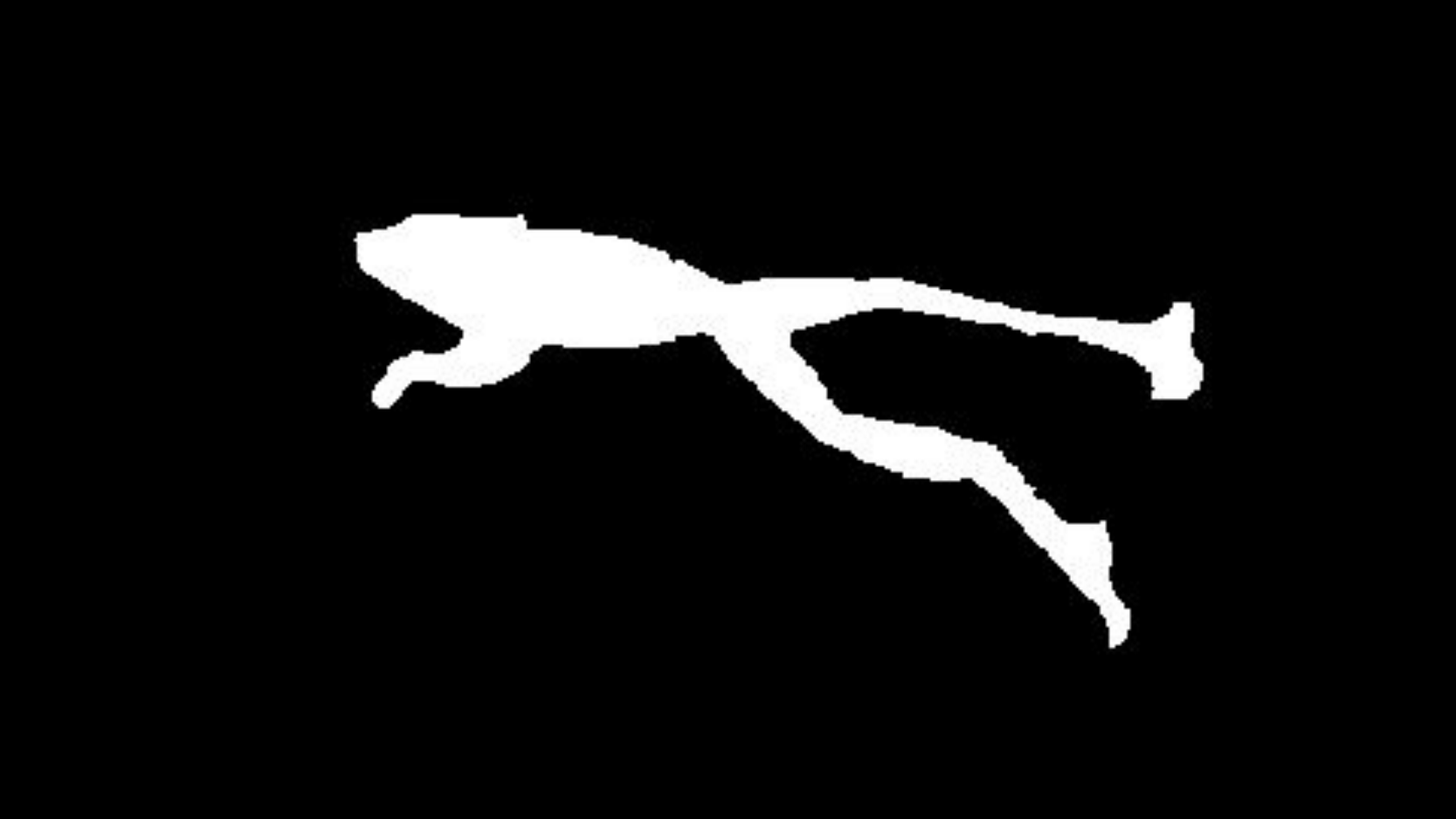}\,&
 \includegraphics[width=0.15\textwidth]{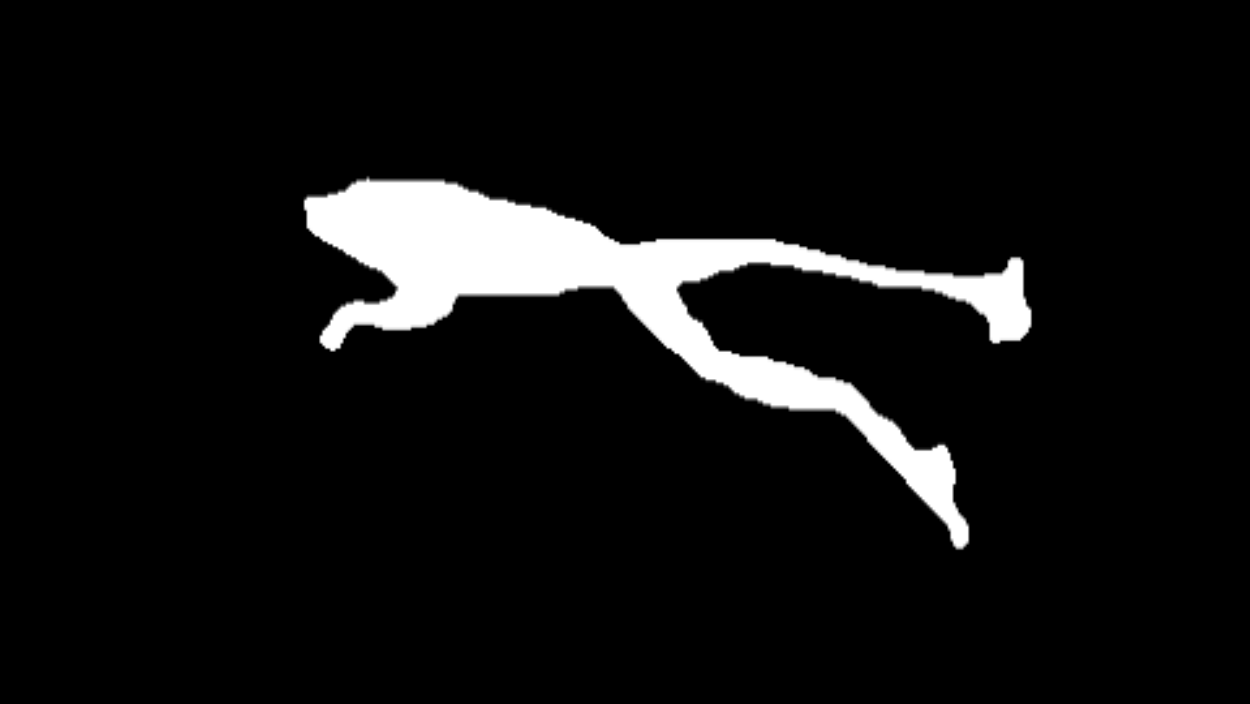}\,&\\

 image & proposed result & ground truth

		\end{tabular}
		\caption{Sample results on the SegTrack v2 benchmark.  
		}
		\label{results_seg_trackv2}
		\end{center}
\end{figure}

\begin{table}
            \centering
            \caption{Results on the SegTrack v2 dataset~\cite{c11}.} 
            
		\begin{tabular}{@{\hspace{0cm}}l @{\hspace{0.1cm}}  | @{\hspace{0.1cm}}   c @{\hspace{0.2cm}} c @{\hspace{0.1cm}}c  @{\hspace{0.2cm}}c @{\hspace{0.2cm}} c @{\hspace{0cm}}}
 \toprule
 Sequence/Object  & SPT+CSI \cite{c11} &   MSK~\cite{c2}&   JOTS \cite{c33}   &    SPT \cite{c11} &     \emph{ours}      \\ \midrule
 Unsupervised  & \checkmark &  $\times$  & $\times$ & 	   \checkmark & \checkmark        \\
 Online & $\times$ & \checkmark & \checkmark & 	   \checkmark & \checkmark        \\
 \midrule
 Mean per object & $65.9$ &	$67.4$ & $71.8$ &	$62.7$ & $65.7$ \\
 Mean per seq. & $71.2$ & - & $72.2$ &  $68.0$ & $68.6$ \\ \midrule
	\end{tabular}
        \label{SegTrack_result}
\end{table}

In Tab.~\ref{davis_bench}, we report the mean (M), recall (R) and decay (D) values for F-measure ($\it{F}$) and Jaccard's index ($\it{J}$) over the DAVIS$_{16}$ dataset splits for the state-of-the-art methods OSVOS-S \cite{c21}, MSK~\cite{c2}, VPN~\cite{jampani2016video}, SIAMMASK~\cite{siammask}, CTN~\cite{CTN}, PLM \cite{c28}, OFL \cite{c18}, BVS \cite{c29}, FCP \cite{c30}, and JMP \cite{c31}, as well as for our approach. The main difference between ours and the competing methods is that we neither learn a model such as OSVOS-S, MSK, VPN, SIAMMASK, CTN or PLM, nor work at super-pixel level and iteratively optimize the optical flow for the achieved segmentation, as is the case for OFL. We only use tracking for segmentation. 
Yet, our method indeed outperforms all remaining non deep-learning based approaches as well as CNN approaches such as PLM~\cite{c28} or SIAMMASK~\cite{siammask}.  
Qualitative results are shown in Fig.~\ref{results_2017}. Visually, the generated segmentation results are appealing and fine details are well captured. 

\subsubsection{Cost Terms from CNN Segmentation}
Here, we evaluate the proposed method as a postprocessing step for state-of-the-art CNN predictions. If our method indeed carries complementary information to the appearance cues learned for example by OSVOS-S~\cite{c21}, we should be able to achieve an improvement. In Tab.~\ref{davis_bench} and Tab.~\ref{dataterm_2017}, we specifically report the results of our method when we use plain CNN predictions from OSVOS-S~\cite{c21} and CINM~\cite{c20} as data terms (costs) (OSVOS-S + \emph{ours} and CINM + \emph{ours}) in the DAVIS$_{16}$ and DAVIS$_{17}$ datasets, respectively. In all settings, segmentation results are improved by our method.   
	
\subsection{Results on SegTrack v2}
	
In the following, we evaluate our proposed model on the SegTrack v2 dataset. In Tab.~\ref{SegTrack_result}, 
we compare our results to the state of the art methods JOTS \cite{c33}, MSK~\cite{c2} and \cite{c11}. 
Li et al.~\cite{c11} propose two variants of their method: 1. the online version Segment Pool Tracking (SPT), 2. the offline version with subsequent refinement of the segments within each frame, i.e. Composite Statistical Inference (CSI). 
Similar to SPT, our method operates in an online fashion and does not require dataset specific training. However, we only use tracking information to compute segments, whereas SPT incrementally trains a global model of the appearance of objects.
Yet, our approach produces results within the range of the top performing methods and improves over SPT. 
Several qualitative results for SegTrack v2 are given in Fig.~\ref{results_seg_trackv2}. Our segmentation is able to capture fine details of the objects such as the slender legs of the frog in the \emph{frog} sequence and the arm and hand of the monkey in the \emph{monkey} sequence. 

\section{Conclusions}
	
We have proposed a variational method for single and multiple object segmentation tracking scenarios. It leverages  
optical flow estimations as well as image boundary estimations for the propagation of labels through video sequences, where a key frame annotation is provided. Deep learning based methods are addressing video object segmentation with high computational complexity and sequence specific training. In contrast, our method only considers the first frame annotations and achieves competitive results without an expensive training procedure. In application scenarios which require optical flow as an input (for example for robot navigation), the computation of video object segmentations with our method comes at very low extra costs.
Our proposed method produces visually appealing segmentations  and preserves fine details on the DAVIS$_{16}$, DAVIS$_{17}$ and SegTrack v2 datasets. 

\section*{Acknowledgment}
We acknowledge funding from the European Research Council (ERC) under the European Union’s Horizon 2020 research and innovation programme (grant agreement No. 741215, ERC Advanced Grant INCOVID), the DFG Research Grant KE2264/1-1, and the NVIDIA Corporation for the donation of a GPU.

\bibliographystyle{IEEEtran}
\bibliography{IEEEabrv,manuscript}

\end{document}